\documentclass[10.5pt, a4paper,english]{article}

 \usepackage{amsmath,amssymb}
 \usepackage{bm}
 \usepackage{ascmac}
 \usepackage{amsthm}
 \usepackage{enumerate}
 \usepackage[dvips]{color}
 \usepackage{here}
\usepackage[pdftex]{graphicx}

\usepackage{mathrsfs} 

\theoremstyle{definition}
\newtheorem{theorem}{Theorem}[section]

\newtheorem{definition}{Definition}[section]

\makeatletter 
 \def\section{\@startsection {section}{1}{\z@}{3.5ex plus -1ex minus -.2ex}{2.3 ex plus .2ex}{\bf}}
 \def\@seccntformat#1{\csname the#1\endcsname.\ }
 \makeatother
 
  \makeatletter 
 \def\subsection{\@startsection {subsection}{1}{\z@}{3.5ex plus -1ex minus -.2ex}{2.3 ex plus .2ex}{\bf}}
 \def\@seccntformat#1{\csname the#1\endcsname.\ }
 \makeatother

\numberwithin{equation}{section} 

\newcommand{\argmin}{\operatornamewithlimits{argmin}}
\newcommand{\argmax}{\operatornamewithlimits{argmax}}

\usepackage{algorithm}
\usepackage{algcompatible}

\newcommand{\1}{\mbox{1}\hspace{-0.25em}\mbox{l}}

\usepackage{slashbox}

\usepackage{geometry}
\begin{document}
\begin{center}
{\Large Bayesian Optimization for Distributionally Robust Chance-constrained Problem} \vspace{5mm} 

Yu Inatsu$^{1,\ast}$ \ \ \ \ Shion Takeno$^1$ \ \ \ \ Masayuki Karasuyama$^1$ \ \ \ \ Ichiro Takeuchi$^{1,2}$ \vspace{3mm}   

$^1$ Department of Computer Science, Nagoya Institute of Technology    \\
$^2$ RIKEN Center for Advanced Intelligence Project \\
$^\ast$ E-mail: inatsu.yu@nitech.ac.jp
\end{center}

\vspace{5mm} 

\begin{abstract}
In black-box function optimization, we need to consider not only controllable design variables but also uncontrollable stochastic environment variables. 
In such cases, it is necessary to solve the optimization problem by taking into account the uncertainty of the environmental variables. 
Chance-constrained (CC) problem, the problem of maximizing the expected value under a certain level of constraint satisfaction probability, is one of the practically important problems in the presence of environmental variables. 
In this study, we consider distributionally robust CC (DRCC) problem and propose a novel DRCC Bayesian optimization method for the case where the distribution of the environmental variables cannot be precisely specified. 
We show that the proposed method can find an arbitrary accurate solution with high probability in a finite number of trials, and confirm the usefulness of the proposed method through numerical experiments. 
\end{abstract}

\section{Introduction}
In this study, we consider a black-box function optimization problem with two types of variables called \emph{design variables} which are fully controllable and \emph{environmental variables} which change randomly depending on the uncertainty of the environment. 
Under the presence of these two types of variables, the goal is to identify the  design variables that optimize the black-box function by taking into account the uncertainty of environmental variables.
In the past few years, Bayesian Optimization (BO) framework that takes the uncertain environmental variables into considerations have been studied in various setups (see \S\ref{subsec:related-work}).
In this paper, we study one of such problems called \emph{distributionally robust chance-constrained (DRCC)} problem. 
The DRCC problem is an instance of constrained optimization problems in an uncertain environment, which is important in a variety of practical problems in science and engineering. 

The goal of a CC problem is to identify the design variables that maximize the expectation of the objective function under the constraint that the probability of the constraint function exceeding a given threshold is greater than a certain level.
Let $f({\bm x},{\bm w} )$ and $g({\bm x},{\bm w} )$ be the unknown objective and constraint functions, respectively, both of which depend on the design variables ${\bm x} \in \mathcal{X}$ and the environmental variables ${\bm w} \in \Omega$.
For a given threshold $h \in \mathbb{R}$ and a level $\alpha \in (0,1)$, the CC problem is formulated as 
\begin{subequations} 
\begin{align}
&\argmax _{ {\bm x} \in \mathcal{X} } \int_\Omega f ({\bm x} ,{\bm w} )  p^\dagger ({\bm w} )  \text{d} {\bm w} \label{eq:mean_f}\\
&\text{subject \ to}\  \int _{\Omega}  \1 [ g({\bm x},{\bm w} ) >h ] p^\dagger ({\bm w} )  \text{d} {\bm w}  >\alpha, \label{eq:ptr_g}
\end{align}
\end{subequations}
where $\1[ \cdot] $ is the indicator function and $p^\dagger ({\bm w} )$ is the probability density function of the environmental variables ${\bm w} $. 
When $p^\dagger ({\bm w} )$ is known, there is a method to solve the CC problem (see \S\ref{subsec:related-work}).

In this study, we consider the case where $p^\dagger ({\bm w})$ is unknown as is commonly encountered in practice.
Here, we formulate the uncertainty of $p^\dagger ({\bm w})$ using a measure called \emph{distributionally robustness}.
Let $\mathcal{A}$ be the user-specified candidate distribution family of ${\bm w}$. 
Then, the DRCC problem is defined as 
\begin{align}
\argmax _{ {\bm x} \in \mathcal{X} } F({\bm x}) \ 
\text{subject \ to}\  G({\bm x})  >\alpha, \label{eq:DRCCO}
\end{align}
where $F({\bm x})$ and $G({\bm x})$ are defined as 
\begin{subequations} 
\begin{align}
F ({\bm x} ) &= \inf _{ p({\bm w} ) \in \mathcal{A}  } \int _{\Omega}  f({\bm x},{\bm w} ) p ({\bm w} )  \text{d} {\bm w} , \label{eq:DRCCO_F} \\
G ({\bm x} ) &= \inf _{ p({\bm w} ) \in \mathcal{A}  } \int _{\Omega}  \1 [ g({\bm x},{\bm w} ) >h ] p ({\bm w} )  \text{d} {\bm w} \label{eq:DRCCO_G}.
\end{align}
\end{subequations}
The solution of this problem is robust with respect to the misspecification of the distributions because $F({\bm x})$ and $G({\bm x})$ are defined by considering the worst case scenario among the candidate distribution families.

For the surrogate models for unknown objective function $f({\bm x},{\bm w} )$ and the constraint function $g({\bm x},{\bm w} )$, we employ Gaussian Process (GP) models and study the above DRCC problem in the context of BO framework.
The main technical challenges in this problem is in the characterization of the posterior distributions of $F ({\bm x} )$ and $G ({\bm x} )$.
In this study, we derive credible intervals of $F ({\bm x} )$ and $G ({\bm x} )$ which can be effectively used for solving the DRCC problem in BO framework.
We call the proposed method \emph{Distributionally Robust Chance-constrained Bayesian Optimization (DRCC-BO)} method. 

\subsection{Related Work}
\label{subsec:related-work}
Black-box function optimization problems using a GP surrogate model \cite{williams2006gaussian} have been extensively studied in the context of BO (see, e.g., \cite{settles2009active,shahriari2016taking}).
The constraint in the form of \eqref{eq:DRCCO} is closely related with level set estimation (LSE) problem in which a GP surrogate model is often employed \cite{bryan2006active,Gotovos:2013:ALL:2540128.2540322,zanette2018robust,inatsu2020-b-active,sui2015safe,turchetta2016safe,sui2018stagewise,wachi2018safe}.
The problem \eqref{eq:DRCCO} is also closely related to constrained BO, which has also been studied extensively in the literature \cite{gardner2014bayesian,hernandez2016general}.
In the past few years, various problem settings concerning uncertain environmental variables have been considered in BO and LSE problems.
The most standard approach to deal with the uncertainty in environmental variables is to consider the expected value of $f(\bm x, \bm w)$ or/and $g(\bm x, \bm w)$.
Fortunately, when the GP is employed as the surrogate model for $f(\bm x, \bm w)$ and $g(\bm x, \bm w)$, its expected value is also represented as a GP, so the acquisition functions (AFs) of BO and LSE problems can be easily constructed. 

On the other hand, in many practical problems, the expected value is often not enough, and other risk measures that can no longer be expressed as GPs, such as variance and tail probability, need to be considered.
The objective function $F(\bm x)$ and the constraint function $G(\bm x)$ of the DRCC problem \eqref{eq:DRCCO} are also difficult to handle because they are also not represented as GPs even when $f(\bm x, \bm w)$ and $g(\bm x, \bm w)$ are GPs.
In the past few years, there have been several studies that consider various risk measures of the uncertainty in the environmental variables $\bm w$ and robustness with respect to $p^\dagger(\bm w)$ \cite{iwazaki2020-b-bayesian,iwazaki2020bayesian,inatsu2020-c-active,bogunovic2018adversarially,pmlr-v139-nguyen21b,nguyen2021optimizing,inatsu-2021}.
In particular, \cite{amri2021sampling} proposed a BO method for the CC problem, but they assumed that the distribution $p^\dagger(\bm w)$ is known.

Distributionally robust optimization (DRO) problems have long been studied in robust optimization community for ordinary optimization problems in which the objective function and the constraint functions are explicitly formulated (in contrast to expensive black-box functions as we consider in this study) \cite{scarf1958min,rahimian2019distributionally}.
DRCC problem with explicitly formulated objective and constraint functions were also studied in \cite{xie2021distributionally,ho2021distributionally}, and they were applied to practical problems called power flow optimization \cite{xie2017distributionally,fang2019adjustable}.
On the other hand, the study of DRO problems for black-box functions with high evaluation cost has only recently started.
In \cite{pmlr-v108-kirschner20a,pmlr-v108-nguyen20a}, BO methods to find the design variable that maximizes $F(\bm x)$ in \eqref{eq:DRCCO_F} was studied in DR setting.
Furthermore, in \cite{inatsu-2021}, a BO method to efficiently identify the design variables which satisfy  $G({\bm x})>\alpha$ for $G({\bm x})$ in \eqref{eq:DRCCO_G} was studied in DR setting.
However, to the best of our knowledge, there is no existing studies that can be used directly in the DRCC-BO problem.

\subsection{Contribution}
The contributions of this paper are as follows:
\begin{itemize}
 \item A BO method for DRCC problem called DRCC-BO method is proposed.
       Specifically, we propose a novel AF for DRCC problem based on the credible intervals of $F(\bm x)$ in \eqref{eq:DRCCO_F} and $G(\bm x)$ in \eqref{eq:DRCCO_G} when GPs are employed as the surrogate models for $f(\bm x, \bm w)$ and $g(\bm x, \bm w)$.

 \item Under mild conditions, we showed that the proposed method can find an arbitrarily accurate solution to the DRCC problem with high probability in a finite number of trials.

\item We also showed that by designing the DRCC problem with an appropriate choice of candidate distribution families, without knowing the true distribution $p^\dagger(\bm w)$ the proposed method can find an arbitrary accurate solution even for the CC problem  (Theorem \ref{thm:seido-CC}).

 \item The performance of the proposed method is confirmed through numerical experiments with synthetic as well as simulator-based functions.

\end{itemize}

\section{Preliminary}
Let $f: \mathcal{X} \times \Omega \to \mathbb{R}$ and $g: \mathcal{X} \times \Omega \to \mathbb{R}$ be the expensive-to-evaluate black-box functions. 
We assume that $\mathcal{X}$ and $\Omega$ are finite sets. 
For each 
$({\bm x}, {\bm w}) \in \mathcal{X} \times \Omega$, the values of 
$f({\bm x},{\bm w})$ and $g({\bm x},{\bm w})$ are observed as $y^{(f)}=f({\bm x},{\bm w}) + \varepsilon_f$ and $y^{(g)}=g({\bm x},{\bm w}) + \varepsilon_g$, where $\varepsilon_f$ and $\varepsilon_g$ are independent Gaussian distributions following 
$\varepsilon_f \sim \mathcal{N} (0,\sigma^2_{f,\text{noise}}), \  \varepsilon_g \sim \mathcal{N} (0,\sigma^2_{g,\text{noise}})   $. 
In this study, we consider the following two cases for the observation of ${\bm w}$:
\begin{description}
\item [Uncontrollable:] For each trial $t$, ${\bm w}$ cannot be controlled, and its realization is generated from the unknown distribution $P^\dagger$.
\item [Simulator:] For each trial $t$, ${\bm w}$ can be chosen arbitrarily.
\end{description}
%
%
Moreover, we consider the following $\mathcal{A} _t$ as a family of candidate distributions for $P^\dagger$: 
$$
\mathcal{A}_t=\{ \text{Probability\ function} \ p({\bm w}) \mid d( p({\bm w} ),p^\ast _t ({\bm w} ) ) \leq \epsilon_t \} ,
$$  
where $p^\ast _t ({\bm w} )$ is a user-specified reference distribution, $d(\cdot,\cdot)$ is a given distance function between distributions, and $\epsilon_t  >0$. 
Then, under a given threshold $h$, the DR expectation function $F_t ({\bm x})$ and DR probability function $G_t ({\bm x})$ are defined  for each ${\bm x} \in \mathcal{X}$ and $t \geq 1$ as 
\begin{align*}
F_t ({\bm x}) &= \inf _{p({\bm w} ) \in \mathcal{A} _t }   \sum_{ {\bm w} \in \Omega }   f({\bm x},{\bm w} )  p({\bm w}) , \\
G_t ({\bm x}) &= \inf _{p({\bm w} ) \in \mathcal{A} _t }   \sum_{ {\bm w} \in \Omega }   \1[g({\bm x},{\bm w} ) >h] p({\bm w}) .
\end{align*}
The objective of this study is to efficiently find the optimal design variable ${\bm x}^\ast_t $ that maximizes $F_t ({\bm x})$ such that $G_t ({\bm x})$ exceeds a given level $\alpha \in (0,1)$.
In other words, ${\bm x}^\ast_t $ satisfies that 
\begin{align}
{\bm x}^\ast_t  = \argmax_{ {\bm x} \in \mathcal{X} } F_t ({\bm x} ) \quad \text{subject \ to } \quad G_t ({\bm x} ) >\alpha. 
\nonumber 
\end{align}
If the optimal solution ${\bm x}^\ast_t$ does not exist, it is formally defined as $F({\bm x}^\ast _t) = \min _{ {\bm x} \in \mathcal{X} }  F_t({\bm x} )$.

\subsection{Gaussian Process}
In this study, we use GPs as the surrogate model for the black-box functions $f$ and $g$. 
First, we assume that the GPs, $\mathcal{G}\mathcal{P}(0, k^{(f)}(  ( {\bm x},{\bm w}  ),  ( {\bm x}^\prime,{\bm w}^\prime  )  )       )$, 
 and 
$\mathcal{G}\mathcal{P}(0, k^{(g)}(  ( {\bm x},{\bm w}  ),  ( {\bm x}^\prime,{\bm w}^\prime  )  )       )$ are prior distributions of 
$f$ and $g$, respectively. 
Here, $k^{(f)}(  ( {\bm x},{\bm w}  ),  ( {\bm x}^\prime,{\bm w}^\prime  )  ) $, and 
$k^{(g)}(  ( {\bm x},{\bm w}  ),  ( {\bm x}^\prime,{\bm w}^\prime  )  ) $ are positive-definite kernels. 
Then, under the given dataset $\{ ({\bm x}_i,{\bm w}_i,y^{(f)}_i ) \}_{i=1}^t$, the posterior distribution of $f$ also follows the GP, and its posterior mean $ \mu^{(f)}_t ({\bm x},{\bm w}) $ and posterior variance $ \sigma^{(f) 2}_t ({\bm x},{\bm w}) $ are given by 
\begin{equation}
\begin{split}
\mu^{(f)}_t ({\bm x},{\bm w}) &= {\bm k}^{(f)} _t ({\bm x},{\bm w} )^\top  ({\bm K}^{(f)}_t + \sigma^2_{f,\text{noise}} {\bm I}_t )^{-1} {\bm y}^{(f)}_t , \\  
\sigma^{(f) 2}_t ({\bm x},{\bm w} )&= k^{(f)}(  ({\bm x},{\bm w}),  ({\bm x},{\bm w}))  -{\bm k}^{(f)}_t ({\bm x},{\bm w})^\top 
({\bm K}^{(f)}_t + \sigma^2_{f,\text{noise}} {\bm I}_t )^{-1} {\bm k}^{(f)} _t ({\bm x},{\bm w}),
\end{split}\label{eq:post_mean_and_variance_f}
\end{equation} 
where ${\bm k}^{(f)}_t ({\bm x},{\bm w} ) $ is a $t$-dimensional vector with $i$th element  
 $ k^{(f)}(({\bm x},{\bm w} ) ,({\bm x}_i,{\bm w}_i)) $, 
${\bm y}^{(f)}_t = (y^{(f)}_1,\ldots , y^{(f)}_t )^\top $, ${\bm I}_t $,  being the 
$t \times t$ identity matrix, and ${\bm K}^{(f)}_t $ is the $t \times t$ matrix with  
$(j,k)$ element  $ k^{(f)}(({\bm x}_j,{\bm w}_j ) ,({\bm x}_k,{\bm w}_k))$. 
As in the case of $f$, under the dataset $\{ ({\bm x}_i,{\bm w}_i,y^{(g)}_i ) \}_{i=1}^{t}$, the posterior distribution of $g$ is also a GP, and its posterior mean $ \mu^{(g)}_t ({\bm x},{\bm w}) $ and posterior variance $ \sigma^{(g) 2}_t ({\bm x},{\bm w}) $ 
can be obtained by using the same formula as 
\eqref{eq:post_mean_and_variance_f}.

\section{Proposed Method}\label{sec:proposed}
In this section, we propose a BO method for efficiently solving the DRCC problem. 
In our setting, because $f({\bm x} ,{\bm w} )$ and $g({\bm x} ,{\bm w} )$ are random functions, 
$F_t ({\bm x} )$ and $G_t ({\bm x} )$ are also random functions. 
Thus, one of the natural BO methods is to use credible intervals of  $F_t ({\bm x} )$ and $G_t ({\bm x} )$. 
Unfortunately, although $f({\bm x} ,{\bm w} )$ and $g({\bm x} ,{\bm w} )$ follow GPs, $F_t ({\bm x} )$ and $G_t ({\bm x} )$ do not follow 
GP. 
Hence, credible intervals of $F_t ({\bm x} )$ and $G_t ({\bm x} )$ cannot be constructed based on the property of Normal distribution. 
In the next section, we describe how to  construct credible intervals based on \cite{pmlr-v108-kirschner20a} and \cite{inatsu-2021}.

\subsection{Credible Interval}
For each input $({\bm {x}},{\bm w}) \in \mathcal{X} \times  \Omega$ and trial $t$, we define a credible interval of 
$f({\bm {x}},{\bm w})$ as $Q^{(f)}_t ({\bm {x}},{\bm w}) =[ l^{(f)}_t ({\bm {x}},{\bm w}), u^{(f)}_t ({\bm {x}},{\bm w})]$. 
Here, the lower bound $l^{(f)}_t ({\bm {x}},{\bm w})$ and upper bound $u^{(f)}_t ({\bm {x}},{\bm w})$ are given by 
\begin{align*}
l^{(f)}_t ({\bm {x}},{\bm w}) = \mu^{(f)}_t ({\bm {x}},{\bm w}) - \beta^{1/2}_{f,t} \sigma^{(f)}_t ({\bm {x}},{\bm w}) , \\
 u^{(f)}_t ({\bm {x}},{\bm w}) = \mu^{(f)}_t ({\bm {x}},{\bm w}) + \beta^{1/2}_{f,t} \sigma^{(f)}_t ({\bm {x}},{\bm w}), 
\end{align*}
where $\beta^{1/2}_{f,t} \geq 0$. 
Similar to the same definition of $Q^{(f)}_t ({\bm {x}},{\bm w})$, 
we define a credible interval of $g({\bm {x}},{\bm w})$ as $Q^{(g)}_t ({\bm {x}},{\bm w}) =[ l^{(g)}_t ({\bm {x}},{\bm w}), u^{(g)}_t ({\bm {x}},{\bm w})]$. 
Furthermore, we construct a credible interval of $\1[g({\bm x},{\bm w} )>h ]$ by using $Q^{(g)}_t ({\bm {x}},{\bm w})$. 
Let $\eta > 0$ be a user-specified overestimation parameter.\footnote{The parameter $\eta$ is necessary to theoretical guarantees. Details are given in Section \ref{sec:THEOREM}.} 
Then, we define the credible interval of 
$\1[g(\bm x, \bm x) > h]$ as 
\begin{align*}
{Q}^{(\1)}_t ({\bm {x}},{\bm w};\eta) \equiv [{l}^{(\1)}_t ({\bm {x}},{\bm w};\eta),{u}^{(\1)}_t ({\bm {x}},{\bm w};\eta)] 
=
\begin{cases}
[1,1] & \text{if} \ l^{(g)}_t ({\bm x},{\bm w} ) >h-\eta, \\
[0,1] &   \text{if} \ l^{(g)}_t ({\bm x},{\bm w} ) \leq h-\eta \ {\tt and} \ u^{(g)}_t ({\bm x},{\bm w} ) >h, \\
[0,0] & \text{if} \ l^{(g)}_t ({\bm x},{\bm w} ) \leq h-\eta \ {\tt and} \ u^{(g)}_t ({\bm x},{\bm w} )  \leq h.
\end{cases}
\end{align*}
Next, using $Q^{(f)}_t ({\bm {x}},{\bm w})$, we define a credible interval of $F_t({\bm x})$ as $Q^{(F_t)}_t ({\bm x}) \equiv [l^{(F_t)}_t ({\bm x}),  u^{(F_t)}_t ({\bm x})      ]$, where $l^{(F_t)}_t ({\bm x} ) $ and $u^{(F_t)}_t ({\bm x} )$ are calculated as  
\begin{equation}
\begin{split}
l^{(F_t)}_t ({\bm x} ) &= \inf _{ p({\bm w} ) \in \mathcal{A} _t } \sum_{ {\bm w} \in \Omega }  {l}^{(f)}_t ({\bm {x}},{\bm w}) p({\bm w} ),   \\
u^{(F_t)}_t ({\bm x} ) &= \inf _{ p({\bm w} ) \in \mathcal{A} _t } \sum_{ {\bm w} \in \Omega }  {u}^{(f)}_t ({\bm {x}},{\bm w}) p({\bm w} ).  
\end{split}
\label{eq:F_upper_lower}
\end{equation}
Note that if the distance function $d(\cdot,\cdot)$ is the 
$L1$ (or $L2$)-norm, \eqref{eq:F_upper_lower} can be formulated as a linear (or second-order cone) programming problem.  
In both cases, optimization solvers exist to easily calculate \eqref{eq:F_upper_lower}. 
Similarly, we define a credible interval of $G_t({\bm x})$ as $Q^{(G_t)}_t ({\bm x};\eta) \equiv [l^{(G_t)}_t ({\bm x} ;\eta),  u^{(G_t)}_t ({\bm x} ;\eta)      ]$, and its lower and upper bounds are given by 
\begin{equation}
\begin{split}
l^{(G_t)}_t ({\bm x} ;\eta) &= \inf _{ p({\bm w} ) \in \mathcal{A}_t } \sum_{ {\bm w} \in \Omega }  {l}^{(\1)}_t ({\bm {x}},{\bm w};\eta) p({\bm w} ),   \\
u^{(G_t)}_t ({\bm x} ;\eta) &= \inf _{ p({\bm w} ) \in \mathcal{A}_t } \sum_{ {\bm w} \in \Omega }  {u}^{(\1)}_t ({\bm {x}},{\bm w};\eta) p({\bm w} ).  
\end{split}
\label{eq:G_upper_lower}
\end{equation}
Moreover, using $Q^{(G_t)}_t ({\bm x};\eta)$, we define an estimated upper  (resp. lower) set $H_t$ (resp. $L_t$) and a potential upper  set $M_t$. 
Let $\xi >0$ be a user-specified accuracy parameter. 
Then, we define $H_t$, $L_t$ and $M_t$ as 
\begin{equation}
\begin{split}
H_t  &= \{ {\bm {x}} \in \mathcal{X} \mid  l^{(G_t)}_t ({\bm {x}};\eta) > \alpha -\xi \},  \\
L_t  &= \{ {\bm {x}} \in \mathcal{X} \mid  l^{(G_t)}_t ({\bm {x}};\eta) \leq  \alpha -\xi \ {\tt and} \  u^{(G_t)}_t ({\bm {x}};\eta)  \leq \alpha \} , \\
M_t  &= \{ {\bm {x}} \in \mathcal{X} \mid  l^{(G_t)}_t ({\bm {x}};\eta) \leq  \alpha -\xi \ {\tt and} \  u^{(G_t)}_t ({\bm {x}};\eta) > \alpha \}.  \\
\end{split}
\nonumber 
\end{equation}
%
%
%
%
%

\subsection{Acquisition Function}
We propose an AF to determine the next evaluation point. 
Our proposed AF is based on the following utility function:
\begin{equation}
\mathbb{E} [ \max \{ F_t ({\bm x} ) - c^{\text{(best)}}_t , 0 \} ] \times \mathbb{P} [ G_t ({\bm x} ) > \alpha ]. \label{eq:utility_DRCCECI}
\end{equation}
The first term is the expected improvement for $F_t ({\bm x} )$, and the second term is the probability that the DR probability function  $ G_t ({\bm x} )$ is greater than $\alpha$. 
In the context of constrained BOs without environmental variables, this utility is known as the expected constrained improvement (ECI) \cite{gardner2014bayesian}. 
Similarly, in the CCBO framework, \cite{amri2021sampling} proposed the expected feasible improvement (EFI) AF using the same utility. 
Unfortunately, in the DRCC setup, both the first and second terms cannot be calculated analytically because $F_t ({\bm x} )$ and $G_t ({\bm x} )$ do not follow GPs.
Moreover, numerical approximation is also expensive because it requires a re-optimization calculation (inf operation) for all generated samples.
For this reason, instead of \eqref{eq:utility_DRCCECI}, we consider a CI-based utility function which mimics \eqref{eq:utility_DRCCECI}. 

First, we define the current best point $c^{\text{(best)}}_t $ at trial $t$ as 
\begin{align}
c^{\text{(best)}}_t =
\left \{ 
\begin{array}{ll}
\max_ { {\bm x} \in  H_t }  l^{(F_t)} _t ({\bm x} ) & \text{if} \ H_t \neq \emptyset , \\
\min_ { {\bm x} \in  M_t }  l^{(F_t)} _t ({\bm x} ) & \text{if} \ H_t = \emptyset \ {\tt and} \ M_t \neq \emptyset , \\
\min_ { {\bm x} \in  \mathcal{X} }  l^{(F_t)} _t ({\bm x} ) & \text{if} \ H_t = \emptyset \ {\tt and} \ M_t = \emptyset  . \\
\end{array}
\right .
  \nonumber
\end{align}
Using this, we define the CI-based improvement $a^{(F_t)}_t ({\bm x})$ for $F_t$  as
$$
a^{(F_t)}_t ({\bm x}) = \max \{  u^{(F_t)}_t ({\bm x} )- c^{\text{(best)}}_t , 0 \}.
$$
Similarly, we define the CI-based probability  $a^{(G_t)}_t ({\bm x})$ for $G_t >\alpha $ as 
\begin{align}
a^{(G_t)}_t ({\bm x}) =
\left \{ 
\begin{array}{ll}
1 & \text{if} \ {\bm x} \in H_t  , \\
\frac{ u^{(G_t)}_t ({\bm x} ;\eta )-(\alpha-\xi)}{u^{(G_t)}_t ({\bm x} ;\eta )-l^{(G_t)}_t ({\bm x} ;\eta )} & \text{if} \ {\bm x} \in M_t , \\
0 & \text{if} \ {\bm x} \in L_t .
\end{array}
\right .
 \nonumber
\end{align}
By combining these, we propose the following AF $a_t ({\bm x} )$: 
$$
a_t ({\bm x} ) = a^{(F_t)}_t ({\bm x}) \times a^{(G_t)}_t ({\bm x}).
$$
Then, the next selected point is evaluated as follows:
\begin{definition}\label{def:1}
The next design variable ${\bm x}_{t+1}$ to evaluate is selected by 
$$
{\bm x}_{t+1} = \argmax_{ {\bm x} \in H_t \cup M_t }  a_t ({\bm x} ).
$$
\end{definition}
On the other hand, unlike the uncontrollable setting, we also need to select ${\bm w}_{t+1}$ in the simulator setting.
One of the reasonable approaches is to focus on large posterior variances at the selected ${\bm x}_{t+1}$.
Thus, we propose the following selection rule to evaluate ${\bm w}$:
\begin{definition}\label{def:w_selection_rule}
The next environmental variable ${\bm w}_{t+1}$ to evaluate is selected by
$$
{\bm w}_{t+1} = \argmax _{ {\bm w} \in \Omega }    \{  \sigma^{(f) 2}_t   ({\bm x}_{t+1},{\bm w} )  +   \sigma^{(g) 2}_t   ({\bm x}_{t+1},{\bm w} ) \}.
$$
\end{definition}

\subsection{Stopping Conditions and DRCC-BO}
We formulate the stopping condition of the proposed algorithm. 
If it is identified that the constraint $G_t ({\bm x} ) >\alpha$ is not satisfied with high confidence for all the design variables ${\bm x}$, then the algorithm should be stopped because there is no solution.
Alternatively, the algorithm should also be stopped if the difference between the conservative maximum for $F_t ({\bm x} )$ in the points that satisfy the constraint and the optimistic maximum for $F_t ({\bm x} )$ in the points that may satisfy the constraint is sufficiently small.
Based on these ideas, we define the following two stopping conditions:
\begin{description}
\item [(S1)] $L_t = \mathcal{X} $. 
\item [(S2)] $ H_t \neq \emptyset$ and  $ \max_{  {\bm x} \in H_t \cup M_t } u^{(F_t ) }_t ({\bm x} ) - \max_{  {\bm x} \in H_t  } l^{(F_t ) }_t ({\bm x} ) < \xi $.
\end{description}
The pseudocode of the proposed method in uncontrollable and simulator settings are given in Algorithm \ref{alg:1}.

\begin{algorithm}[t]
    \caption{DRCC-BO: BO for DRCC problem}
    \label{alg:1}
    \begin{algorithmic}
        \REQUIRE GP priors $\mathcal{GP}(0,\ k^{(f)}), \ \mathcal{GP}(0,\ k^{(g)})$, threshold $h \in \mathbb{R}$, level $\alpha \in (0,1)$, overestimation parameter $\eta > 0$, $\{\beta_{f,t}\}_{t \geq 1}, \{\beta_{g,t}\}_{t \geq 1}$, accuracy parameter $\xi >0$, $\mathcal{A}_t$ 
        \STATE $H_0\leftarrow \emptyset$, $L_0 \leftarrow \emptyset$, $M_0 \leftarrow \mathcal{X}$, $t \leftarrow 1$
        \WHILE{Both (S1) and (S2) are not satisfied}
           
            \STATE Compute $Q^{(F_t)}_t ({\bm x} )$ and $Q^{(G_t)}_t ({\bm x};\eta )$ for any $\bm{x} \in \mathcal{X}$
            \STATE Select the next evaluation point $\bm{x}_t$
          
	\IF{Uncontrollable setting}
    \STATE Generate the next evaluation point $\bm{w}_t \sim P^\dagger$ 
    \ELSE
    \STATE Select the next evaluation point $\bm{w}_t$
    \ENDIF
 
            \STATE Observe $y^{(f)}_t = f(\bm{x}_t, \bm{w}_t) + \varepsilon_{f,t}$ and $y^{(g)}_t = g(\bm{x}_t, \bm{w}_t) + \varepsilon_{g,t}$
            \STATE Update the GPs by adding  observations, and compute ${H}_t, {L}_t,{M}_t$
            \STATE $t \leftarrow t + 1$
        \ENDWHILE
        \ENSURE No solution exists if (S1) is satisfied, and otherwise $\hat{\bm x}_t = \argmax_{ {\bm x} \in H_{t-1} } l^{(F_t)}_{t-1} ({\bm x} ) $
    \end{algorithmic}
\end{algorithm}

\section{Theoretical Analysis}\label{sec:THEOREM}
In this section, we show the theoretical guarantee on the accuracy and convergence in our proposed algorithm.
First, we assume that the true black-box functions $f$ and $g$ follow GPs $\mathcal{G}\mathcal{P} (0,k^{(f)}(({\bm x},{\bm w}),({\bm x}^\prime,{\bm w}^\prime)))$, and $\mathcal{G}\mathcal{P} (0,k^{(g)}(({\bm x},{\bm w}),({\bm x}^\prime,{\bm w}^\prime)))$, respectively.  
Moreover, as a technical condition, we assume that the posterior variances 
 $k^{(f)}(({\bm x},{\bm w}),({\bm x},{\bm w})) \equiv \sigma^{(f)2}_0 ({\bm x},{\bm w} ) $ and 
 $k^{(g)}(({\bm x},{\bm w}),({\bm x},{\bm w})) \equiv \sigma^{(g)2}_0 ({\bm x},{\bm w} ) $ satisfy 
\begin{align*}
&\max_{( {\bm x},{\bm w} ) \in \mathcal{X}\times \Omega } \sigma^{(f)2}_0 ({\bm x},{\bm w} )  \leq 1,   \  \max_{( {\bm x},{\bm w} ) \in \mathcal{X}\times \Omega } \sigma^{(g)2}_0 ({\bm x},{\bm w} ) \leq 1,    \\
&0< \sigma^{(g)2}_{0,min}  \equiv \min_{( {\bm x},{\bm w} ) \in \mathcal{X}\times \Omega } \sigma^{(g)2}_0 ({\bm x},{\bm w} ) .
\end{align*}
Here, let $\kappa^{(f)}_T,\ \kappa^{(g)}_T$ be the maximum information gain of $f$ and $g$ at trial $T$, respectively. 
Note that the maximum information gain is a commonly used complexity measure in the context of the GP-based BO method 
(see, e.g.,  \cite{SrinivasGPUCB}). 
The value $\kappa^{(f)}_T $ can be expressed as 
$
\kappa^{(f)}_T = \max_{ S \subset \mathcal{X}\times \Omega } I({\bm y}^{(f)}_S;f ),
$
where $I({\bm y}^{(f)};f)$ is the mutual information between ${\bm  y}^{(f)}$ and $f$.
Similarly, $\kappa^{(g)}_T$ can be expressed by using $I({\bm y}^{(g)};g)$. 
Next, we define the goodness of the estimated $\hat{\bm x}_t$ as follows:
\begin{definition}\label{def:C-accuracy}
For a given positive constant $C$, we define the estimated solution $\hat{\bm x}_t$ as the $C$-accurate solution if $\hat{\bm x}_t$ satisfies the following inequalities:
$$
F_t ({\bm x}^\ast_t ) - F_t (\hat{\bm x}_t ) < C, \quad G_t  (\hat{\bm x}_t ) > \alpha - C.
$$
\end{definition}
Then, the following theorem holds:
\begin{theorem}\label{thm:seido}
Let $h \in \mathbb{R}$, $\alpha \in (0,1)$, $t \geq1$, $\delta \in (0,1)$, and define 
$\beta_{f,t} = \beta_{g,t} = 2 \log (2|\mathcal{X}\times \Omega | \pi^2 t^2 /(3\delta ) ) \equiv \beta_t$. 
For a user-specified accuracy parameter $\xi >0 $, define an overestimation parameter $\eta >0$ as 
$$
\eta = \min \left \{
\frac{\xi \sigma^{(g)}_{0,min} }{2}, \frac{\xi ^2 \delta \sigma^{(g)}_{0,min}}{8 |\mathcal{X}\times \Omega | }
\right \}. 
$$
Then, when Algorithm \ref{alg:1} is performed, with a probability of at least $1-\delta$, the following 
 holds for any $t$ and $\mathcal{A}_t$:
\begin{itemize}
\item If (S1) is satisfied, then $G_t ({\bm x} )  \leq \alpha $ for all ${\bm x} \in \mathcal{X} $, that is, the DRCC problem has no solution.
\item If (S2) is satisfied, then $\hat{\bm x} _t $ is the $2 \xi$-accurate solution.
\end{itemize}
Moreover, these results do not depend on whether the simulator or uncontrollable setting is used.
\end{theorem}
We would like to note that although Theorem \ref{thm:seido} guarantees the returned solution by Algorithm \ref{alg:1} is good, but does not state whether the stopping conditions are satisfied or not.
The sufficient conditions for stopping conditions to be satisfied need to be considered for the simulator and uncontrollable settings, separately.
First, we give the sufficient conditions in the simulator setting.
\begin{theorem}\label{thm:convergence1}
Under the same condition as in Theorem \ref{thm:seido}, let 
$T$ be the smallest positive integer satisfying the following inequalities:
\begin{align}
 \frac{    \beta_{T^2}   (    C_{1,f}  \kappa^{(f)}_T    + C_{1,g}  \kappa^{(g)}_T  )             }{T} <  \min \left \{ \frac{\xi^4}{4}    , \frac{\eta^2}{4}    \right \}  ,               \label{eq:mi_ineq}
\end{align}
where  $C_{1,f} = 2/ \log (1+\sigma^{-2}_{f,\text{noise} } ) $ and $C_{1,g} = 2 / \log (1+\sigma^{-2}_{g,\text{noise} } )$. 
Then, under the simulator setting, Algorithm \ref{alg:1} terminates after at most $T^2$ trials.
\end{theorem}

Next, we give the sufficient conditions in the uncontrollable setting. 
In the simulator setting, we can choose any ${\bm w}$ and thus the uncertainty of $f$ and $g$ can be reduced sufficiently. 
In contrast, in the uncontrollable setting, we cannot choose ${\bm w}$, freely.
For this reason, it is desirable to be able to make the uncertainty of all points small in probability.
However, if $p^\dagger ({\bm w} ) = 0$ for some ${\bm w} $, the uncertainty at points including this point is not reduced sufficiently.
To avoid this problem, in the uncontrollable setting, we assume that the true distribution satisfies 
$$
\min _{ {\bm w} \in \Omega } p^\dagger ({\bm w} ) \equiv p_{\text{min}}    >0 . 
$$
Then, the following theorem holds:
\begin{theorem}\label{thm:convergence2}
Under the same condition as in Theorem \ref{thm:seido}, assume that $p_{\text{min}}    >0 $.
Let  
$T$ be the smallest positive integer satisfying the following inequalities:
\begin{equation}
 \frac{    \beta_{T^2}   (    C_{2,f}  \kappa^{(f)}_T    + C_{2,g}  \kappa^{(g)}_T +C_3 )             }{T} <  \min \left \{ \frac{\xi^4}{4}    , \frac{\eta^2}{4}    \right \} , \label{eq:mi_ineq2}
\end{equation}
where  $C_{2,f} = (4    p^{-1}_{\text{min}}     )/ \log (1+\sigma^{-2}_{f,\text{noise} } ) $, $C_{2,g} = (4  p^{-1}_{\text{min}} ) / \log (1+\sigma^{-2}_{g,\text{noise} } )$ and $C_3 = 16  p^{-1}_{\text{min}} \log (5/\delta ) $.  
Then, under the uncontrollable setting, with a probability of at least $1-\delta$,  Algorithm \ref{alg:1} terminates after at most $T^2$ trials.
\end{theorem}
Furthermore, we give a theorem that it is possible to link the DRCC problem to the CC problem by choosing $\mathcal{A}_t$ appropriately. 
Specifically, it ensures that the solution to the DRCC problem is also a good solution to the CC problem.
Here, we consider the following CC problem 
\begin{align*}
\tilde{F} ({\bm x}) &=    \sum_{ {\bm w} \in \Omega }   f({\bm x},{\bm w} )  p^\dagger ({\bm w}) , \\
\tilde{G} ({\bm x}) &=   \sum_{ {\bm w} \in \Omega }   \1[g({\bm x},{\bm w} ) >h] p^\dagger ({\bm w}) , \\
\tilde{\bm x}^\ast  &= \argmax_{ {\bm x} \in \mathcal{X} } \tilde{F} ({\bm x} ) \quad \text{subject \ to } \quad \tilde{G} ({\bm x} ) >\alpha, 
\end{align*}
where 
if the optimal solution $\tilde{\bm x}^\ast $ does not exist, it is formally defined as $\tilde{F}(\tilde{\bm x}^\ast ) = \min _{ {\bm x} \in \mathcal{X} }  \tilde{F} ({\bm x} )$.
As with the DRCC problem, we define the goodness of the solution to the CC problem as follows:
\begin{definition}\label{def:C-accuracy-CC}
For a given positive constant $C$, we define the solution $\hat{\bm x}_t$ as the $C$-accurate solution to the CC problem if $\hat{\bm x}_t$ satisfies the following inequalities:
$$
\tilde{F} (\tilde{\bm x}^\ast ) - \tilde{F} (\hat{\bm x}_t ) < C, \quad \tilde{G}  (\hat{\bm x}_t ) > \alpha - C.
$$
\end{definition}
Then, the following theorem holds:
\begin{theorem}\label{thm:seido-CC}
Under the uncontrollable setting, let $h \in \mathbb{R}$, $\alpha \in (0,1)$, $t \geq1$, $\delta \in (0,1)$, and define 
$\beta_{f,t} = \beta_{g,t} = 2 \log (2|\mathcal{X}\times \Omega | \pi^2 t^2 /(3\delta ) )\equiv \beta_t$. 
For a user-specified accuracy parameter $\xi >0 $, define $\alpha^\prime =\alpha - \xi$ and an overestimation parameter $\eta >0$ as 
$$
\eta = \min \left \{
\frac{\xi \sigma^{(g)}_{0,min} }{2}, \frac{\xi ^2 \delta \sigma^{(g)}_{0,min}}{8 |\mathcal{X}\times \Omega | }
\right \}. 
$$
Furthermore, let $p^\ast_t ({\bm w})$ be an empirical distribution of ${\bm w}$, and let 
\begin{align*}
\epsilon_t &= |\Omega | \sqrt{\frac{1}{2t} \log \left ( \frac{ |\Omega| \pi^2 t^2 }{3 \delta}   \right )}, \\ 
d( p_1 ({\bm w} ), p_2 ({\bm w} ) ) &= \sum _{  {\bm w} \in \Omega }   |  p_1 ({\bm w} )- p_2 ({\bm w} ) |. 
\end{align*}
Then, when Algorithm \ref{alg:1} is performed by using $\alpha^\prime$, with a probability of at least $1-2\delta$, the following  
holds for any $t \geq {T}$:
\begin{itemize}
\item If (S1) is satisfied, then $\tilde{G} ({\bm x} )  \leq \alpha $ for all ${\bm x} \in \mathcal{X} $, that is, the CC problem has no solution,
\item If (S2) is satisfied, then $\hat{\bm x} _t $ is the $3 \xi$-accurate solution for the CC problem,
\end{itemize}
where $T$ is the smallest positive integer satisfying 
$$
^\forall n \geq T, 2(1+\beta^{1/2}_{f,1} ) \epsilon _n < \xi.
$$
\end{theorem}
Finally, the results of the theorems obtained in this section are summarized in Table \ref{tab:theorem_summary}.
\begin{table*}[htb]
  \begin{center}
    \caption{Probabilities of accuracy and algorithm termination in each setting}
\scalebox{1.0}{
    \begin{tabular}{c||ccc} \hline \hline
       &Simulator  & Uncontrollable &   Uncontrollable using Theorem \ref{thm:seido-CC} setting   \\ \hline 
     $2\xi$-accuracy for DRCC                             &  $1-\delta$ &  $1-\delta$ & $1-\delta$ \\
     $3\xi$-accuracy for CC                                 &  NA & NA & $1-2\delta$ \\
     Algorithm termination                                     &  $1$ & $1-\delta$ & $1-\delta$ \\
      $2\xi$-accuracy for DRCC and termination      &  $1-\delta$ & $1-2\delta$ & $1-2\delta$ \\
     $3\xi$-accuracy for CC and termination           &  NA & NA & $1-3\delta$   \\  \hline  \hline
    \end{tabular}
}
    \label{tab:theorem_summary}
  \end{center}
\end{table*}

Note that the order of maximum information gains $\kappa^{(f)}_T$ and $\kappa^{(g)}_T$ is known to be sublinear under mild conditions 
 \cite{SrinivasGPUCB}. 
Therefore, noting that the order of 
$\beta_{f,T} =\beta_{g,T}$ is $O ( \log T )$, the positive integer $T$ satisfying \eqref{eq:mi_ineq} and \eqref{eq:mi_ineq2}  exists.  

\section{Numerical Experiments}\label{sec:sec5}
In this section, we confirm the performance of the proposed method in simulator and uncontrollable settings using synthetic functions and real-world simulations. 
In this experiment, both design and environment variables were set to one dimension, and the following Gaussian kernels were used as the kernel functions:
\begin{align*}
k^{(f)}( (x,w),(x^\prime,w^\prime ) ) &= \sigma^2_{f,\text{ker}} \exp (-\| {\bm \theta}-{\bm \theta}^\prime \|^2_2 /L_f), \\
k^{(g)}( (x,w),(x^\prime,w^\prime ) ) &= \sigma^2_{g,\text{ker}} \exp (-\| {\bm \theta}-{\bm \theta}^\prime \|^2_2 /L_g),
\end{align*}
where ${\bm \theta } = (x,w) $.
We used the $L1$-norm as the distance between distributions, and set $\epsilon_t = 0.15$. 
Here, for simplicity, we set the overestimation parameter $\eta$ to 0 and the accuracy parameter to $\xi = 10^{-12}$. 
In all experiments, we evaluated the performance of each algorithm using the following utility gap $\text{UG}_t $:
\begin{align*}
{\rm UG}_t 
= \left \{
\begin{array}{ll}
F_t(x^\ast _t ) -  F _t ( \hat{x} _t )    & \text{if} \ H_t \neq \emptyset \ {\tt and } \ G_t (\hat{x}_t ) >\alpha , \\
F _t (x^\ast _t ) - \min_{x \in \mathcal{X} }  F_t (x) & \text{otherwise},
\end{array}
\right . 
\end{align*}
where $\hat{x}_t$ is given by $\hat{x}_t = \argmax _{x \in H_t} l^{(F_t)}_t (x )$. 
The details of the experimental setting, which are not included in the main body, are described in Appendix \ref{app:exp_detail}.

\subsection{Synthetic Function}\label{syn_experiment}
We evaluate the performance of the proposed method using a synthetic function. 
We used the input space $\mathcal{X}\times \Omega$ as a set of grid points divided by $[-10,10] \times [-10,10]$ into $50 \times 50$ equally spaced. 
Moreover, we used the following black-box functions $f$ and $g$:
\begin{align*}
&f(x,w) =   \exp  (-{x^2}/{4}  ) + 0.6 \exp  (-{(x-8)^2} / {3}  ) 
 +0.3 \exp  (-{(x+9)^2} / {5} ) \\ 
&\quad \quad  \quad \quad + \exp (-{w^2} /{4}) + 0.6 \exp (-{(w-8)^2} / {3} ) +0.3 \exp (-{(w+9)^2} / {5} ) , \\
&g(x,w) = 0.26 (x^2+w^2) - 0.48 xw. 
\end{align*}
In this experiment, we performed a total of three different experiments, one with the simulator setting and two with the uncontrollable setting:
\begin{description}
\item [Simulator:] Under the simulator setting, $p^\ast_t ({\bm w} ) =1/50$ was used as the reference distribution.
\item [Fixed:] Under the uncontrollable setting, the mixture normal distribution $0.5 \mathcal{N} (-5,10) +0.5 \mathcal{N} (5,10) $ discretized on $\Omega$ was used as the true distribution $p^\dagger ({\bm w} )$. 
The reference distribution was set to  $p^\ast_t ({\bm w}) =1/50$. 
\item [Data-driven:] Under the uncontrollable setting, for the true distribution $p^\dagger ({\bm w})$ we used the same as {\sf Fixed}, and for the reference distribution we used the empirical distribution function of ${\bm w}$.
\end{description}
We compared the following six methods:
\begin{description}
			\item [Random:] Select $(x_{t+1},w_{t+1} )$ randomly. 
			\item [US:] Select $(x_{t+1},w_{t+1} )$ by maximizing  the maximum posterior variance of $f$ and $g$, i.e., $(x_{t+1},w_{t+1} )$ is given by 
$$
 (x_{t+1},w_{t+1} ) = \argmax _{ (x,w) \in \mathcal{X} \times \Omega} {\rm US}_t (x,w),
$$
where $ {\rm US}_t (x,w) = \max \{ \sigma^{(f)2}_t (x,w)  ,  \sigma^{(g)2}_t (x,w)    \}   $.
			\item [DRBO:] Use the DRBO  method proposed by \cite{pmlr-v108-kirschner20a}, i.e., $x_{t+1} $ and $w_{t+1}$ are selected by $x_{t+1} = \argmax _{ x \in \mathcal{X} } u^{(F_t)}_t (x) $ and 
$w_{t+1} = \argmax _{ w \in \Omega }  \sigma^{(f) 2} _t (x_{t+1},w) $. 
			\item [DRPTR:] Use the DRPTR  method proposed by \cite{inatsu-2021}, i.e., $(x_{t+1},w_{t+1} )$ is selected  by $(x_{t+1},w_{t+1} ) = \argmax _{ (x,w) \in \mathcal{X} \times \Omega} a^{(2)}_t ({\bm x},{\bm w})    $, where $a^{(2)}_t ({\bm x},{\bm w})$ is given by 
Definition 3.2 of \cite{inatsu-2021}. 
\item [CCBO:] Use the CCBO  method proposed by \cite{amri2021sampling}, i.e.,  $x_{t+1} $ and $w_{t+1}$ are selected by $x_{t+1} = \argmax _{ x \in \mathcal{X} } {\rm EFI} (x)$ and  
$w_{t+1} = \argmax _{ w \in \Omega }  S (x_{t+1},w) $, where ${\rm EFI} (x)$ and $S(x,w)$ are given by (7) and (13) of \cite{amri2021sampling}. 
			\item [Proposed:]  Use Definition \ref{def:1}-\ref{def:w_selection_rule}.
		\end{description}
In the case of uncontrollable setting, we selected only $x_{t+1}$. 
On the other hand, because {\sf US} and {\sf DRPTR} select $x$ and $w$ simultaneously, we modified them in the uncontrollable setting as follows:
\begin{description}
\item [US:] $x_{t+1} = \argmax _{ x \in \mathcal{X} } \mathbb{E}_w [  {\rm US}_t (x,w) ].$
\item [DRPTR:] $x_{t+1} = \argmax _{ x \in \mathcal{X} } \mathbb{E} _w [ a^{(2)}_t ({ x},{ w}) ]$.
\end{description}
Here, the expectation is taken with respect to the empirical distribution of $w$. 
We would like to emphasize that {\sf DRBO} focuses only on the maximization of $F_t ({\bm x})$, and does not consider whether the constraints are satisfied or not.
In contrast, {\sf DRPTR} focuses only on the identification of variables that satisfy the constraints and does not consider the maximization of $F_t ({\bm x})$. 
As for {\sf CCBO}, it is the BO method for the CC problem \eqref{eq:mean_f}--\eqref{eq:ptr_g}, and does not consider the distributionally robustness.

\begin{figure*}[!t]
\begin{center}
 \begin{tabular}{ccc}
 \includegraphics[width=0.33\textwidth]{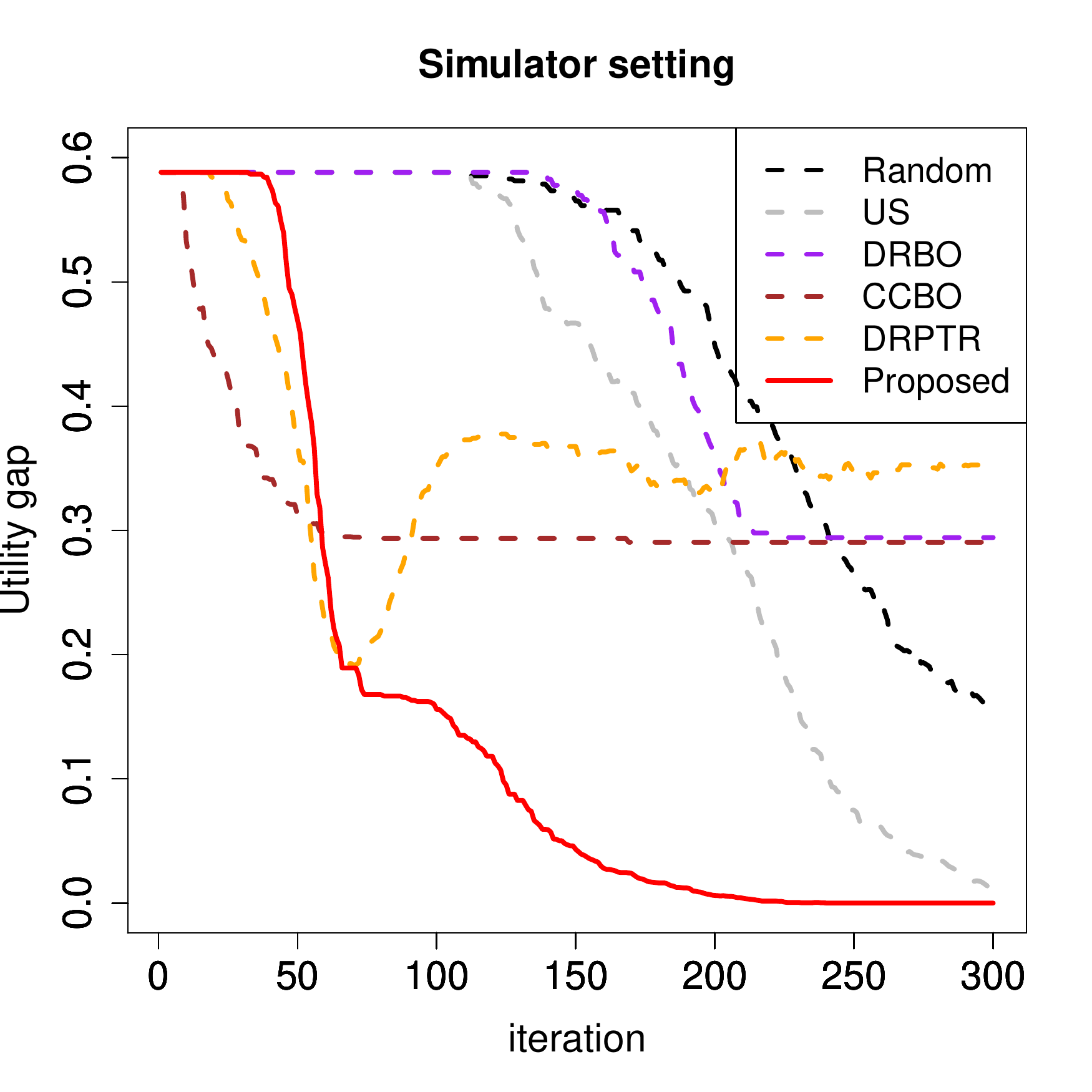} &
 \includegraphics[width=0.33\textwidth]{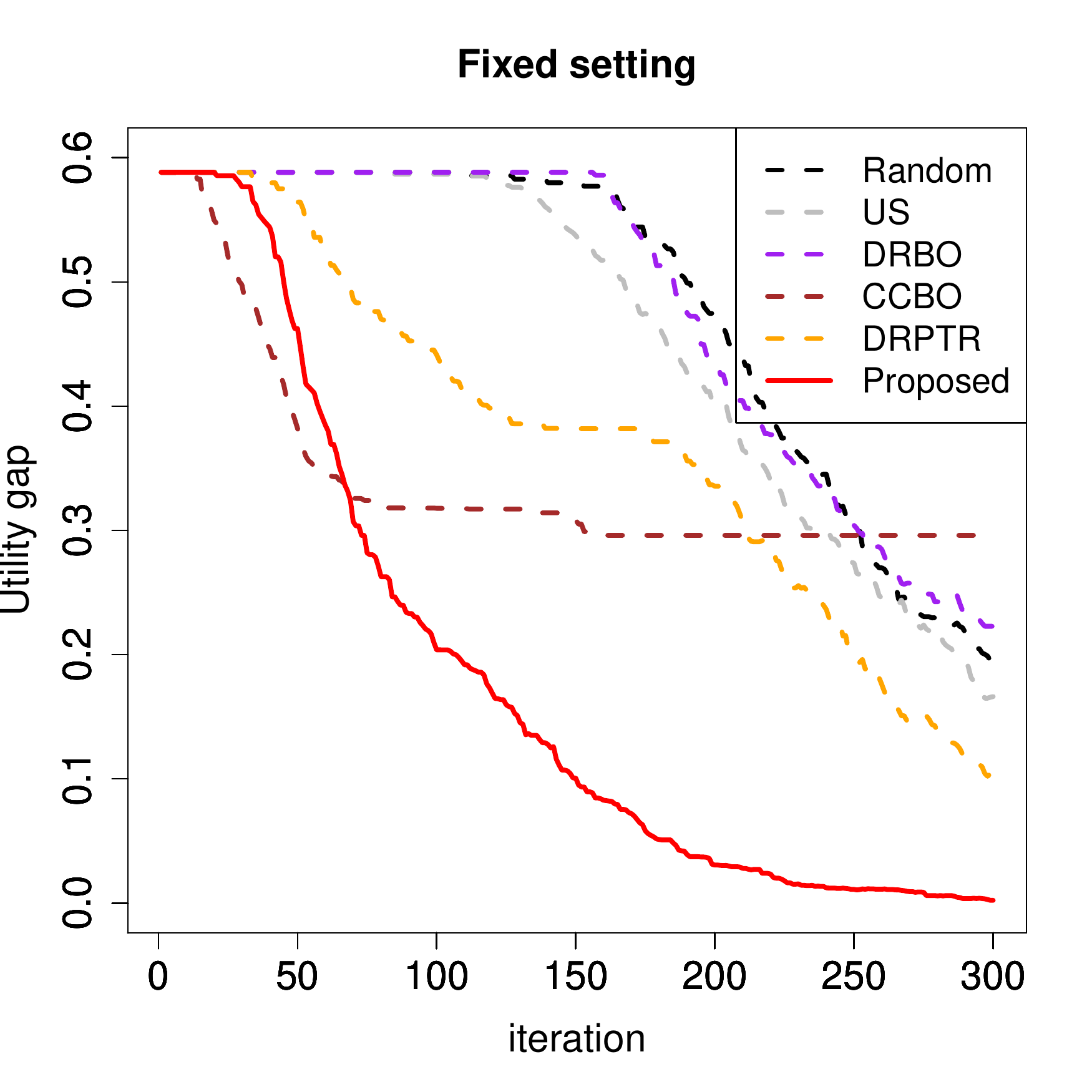} &
 \includegraphics[width=0.33\textwidth]{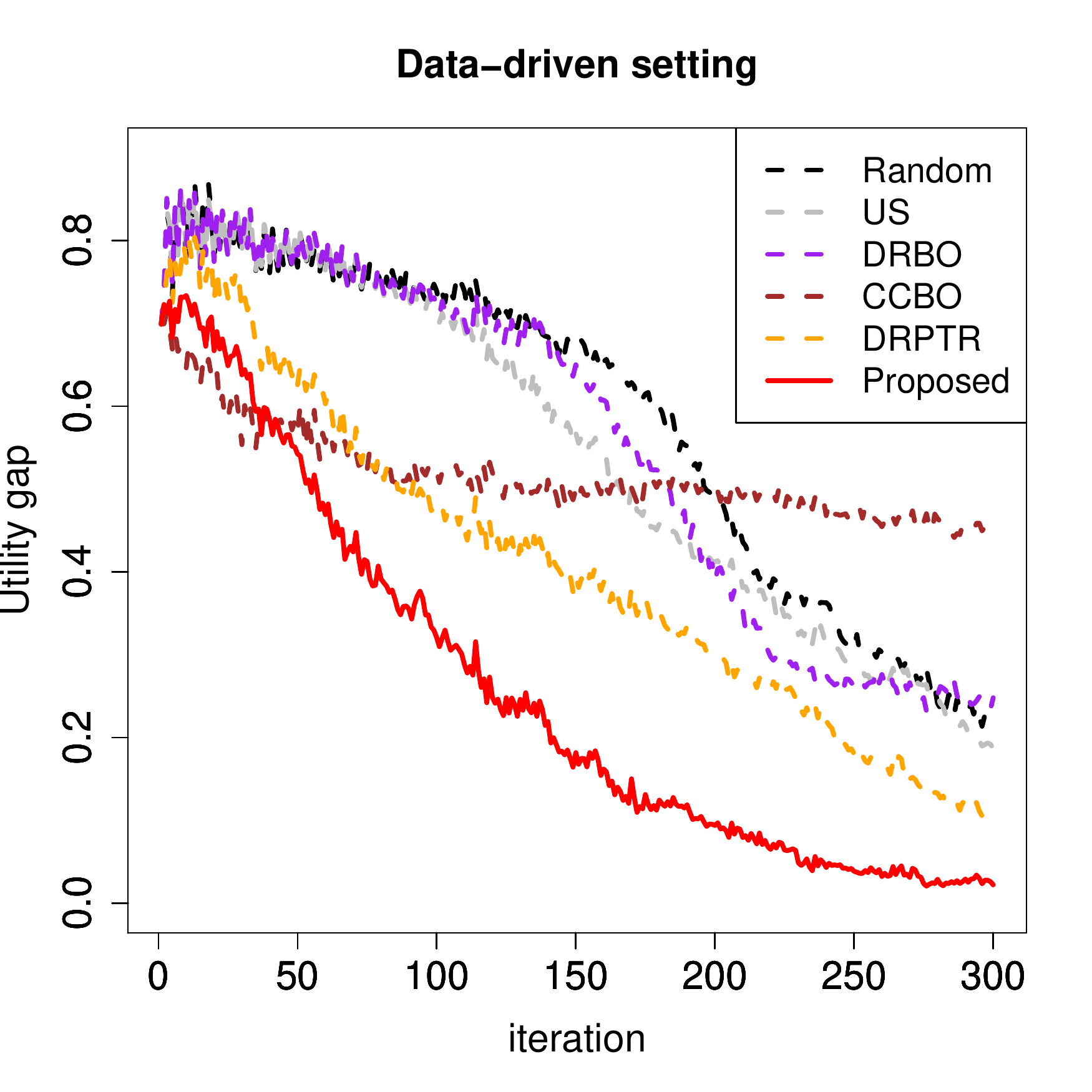} 
 \end{tabular}
\end{center}
 \caption{Average utility gap for each method in simulator and uncontrollable settings.}
\label{fig:exp1}
\end{figure*}

With this setup, we took one initial point at random and ran the algorithms until the number of iterations reached 300.
The simulation was repeated 100 times and the average value of the utility gap at each iteration was calculated. 
From Figure\ref{fig:exp1}, it can be confirmed that the proposed method shows high performance.

\begin{figure*}[!t]
\begin{center}
 \begin{tabular}{cccc}
 \includegraphics[width=0.245\textwidth]{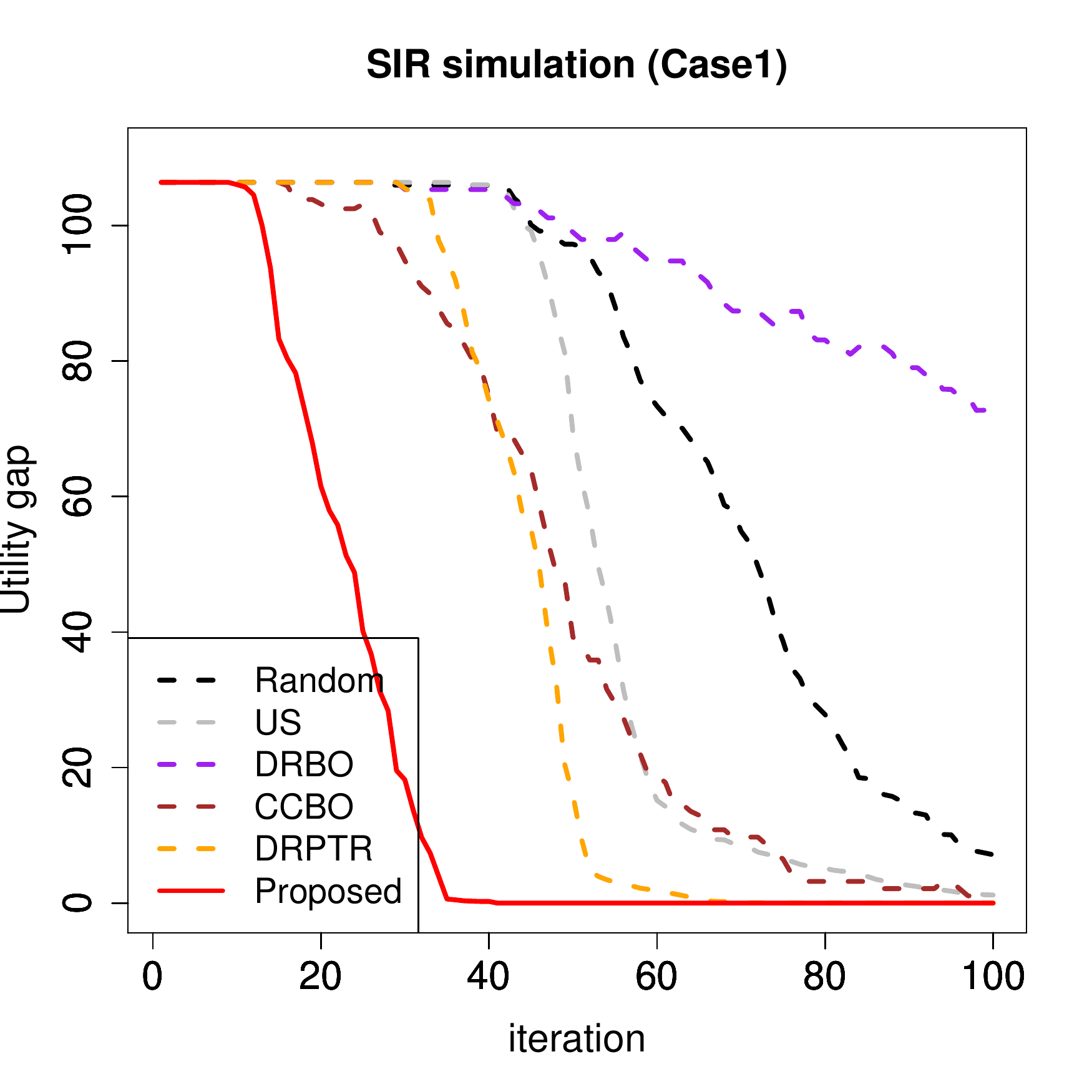} &
 \includegraphics[width=0.245\textwidth]{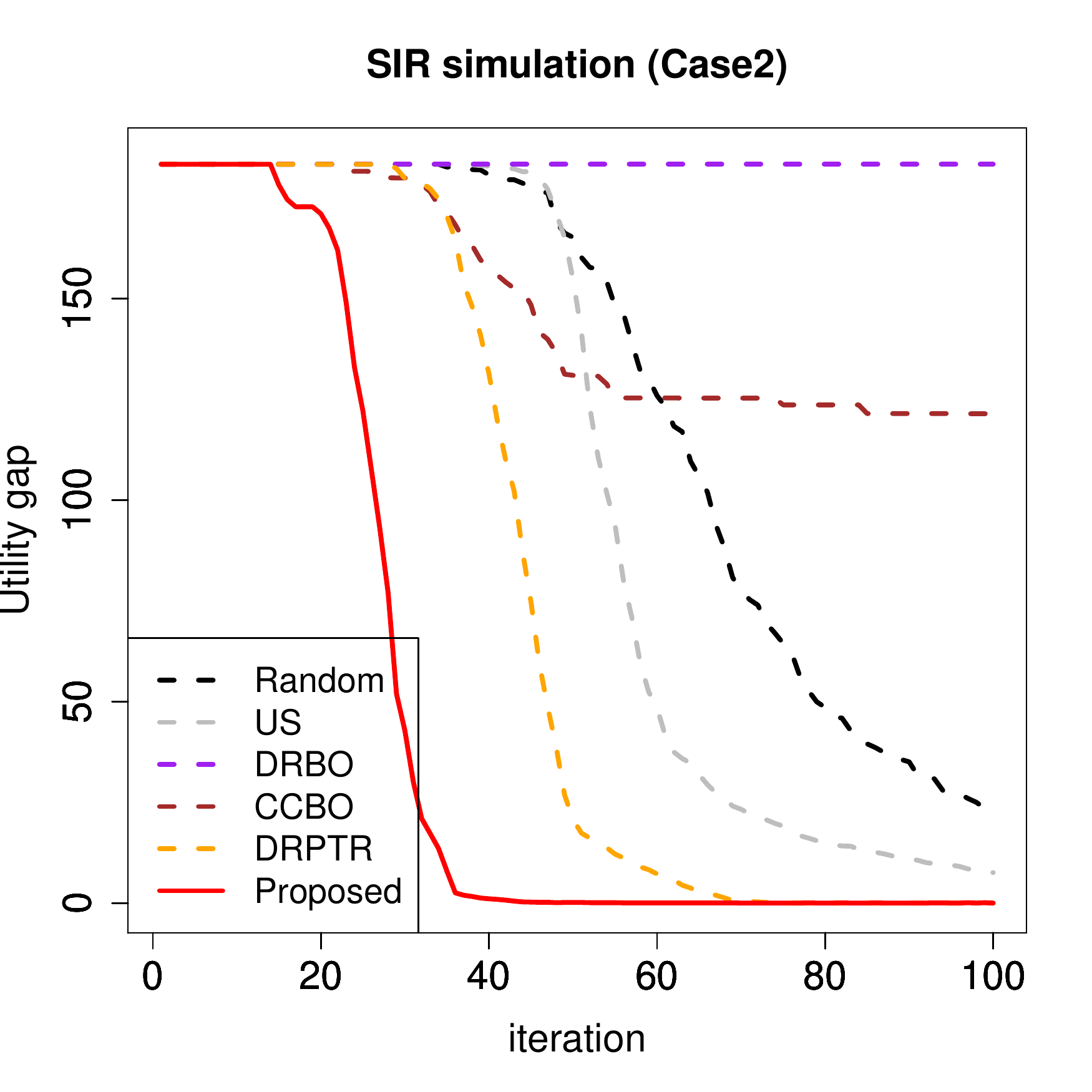} &
 \includegraphics[width=0.245\textwidth]{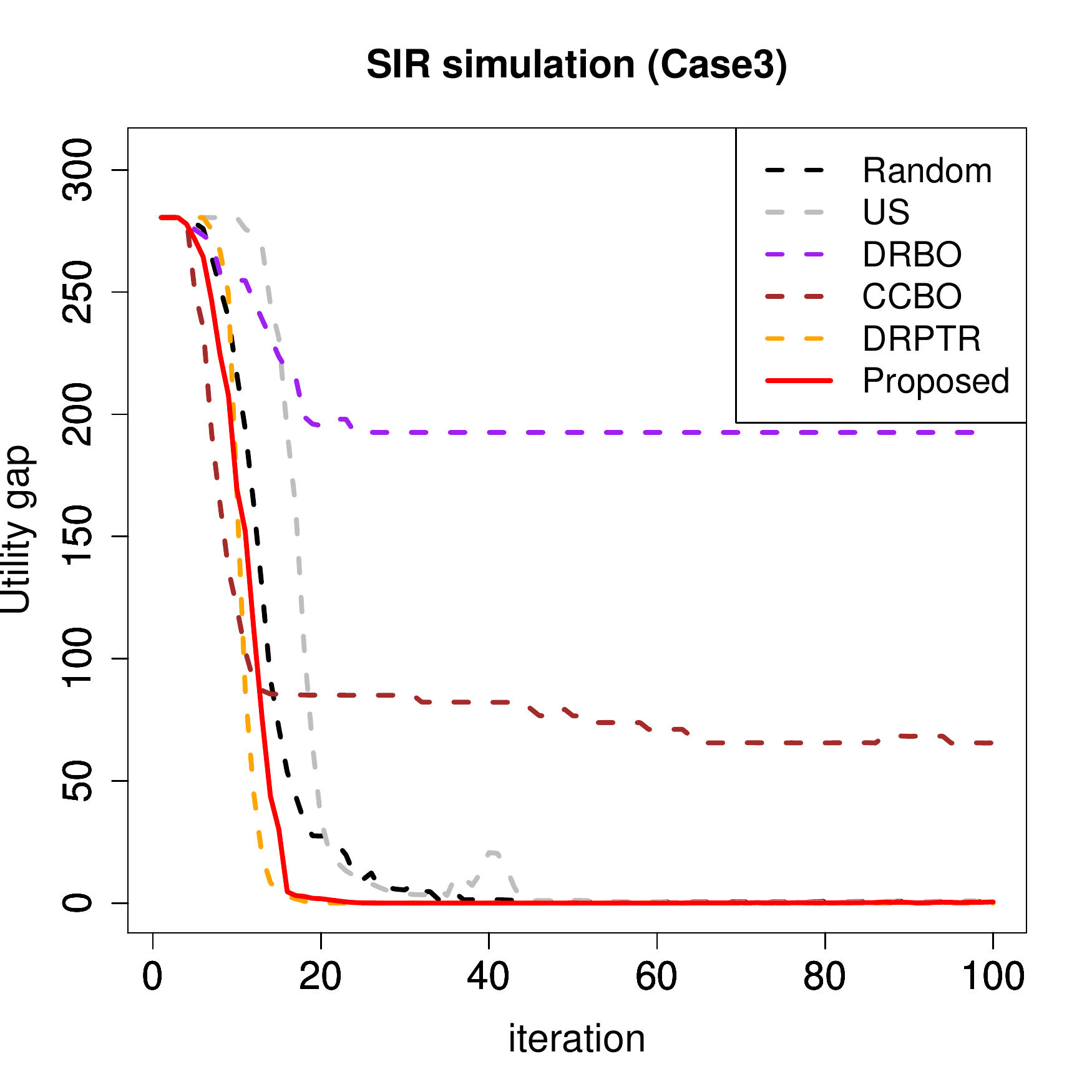} &
 \includegraphics[width=0.245\textwidth]{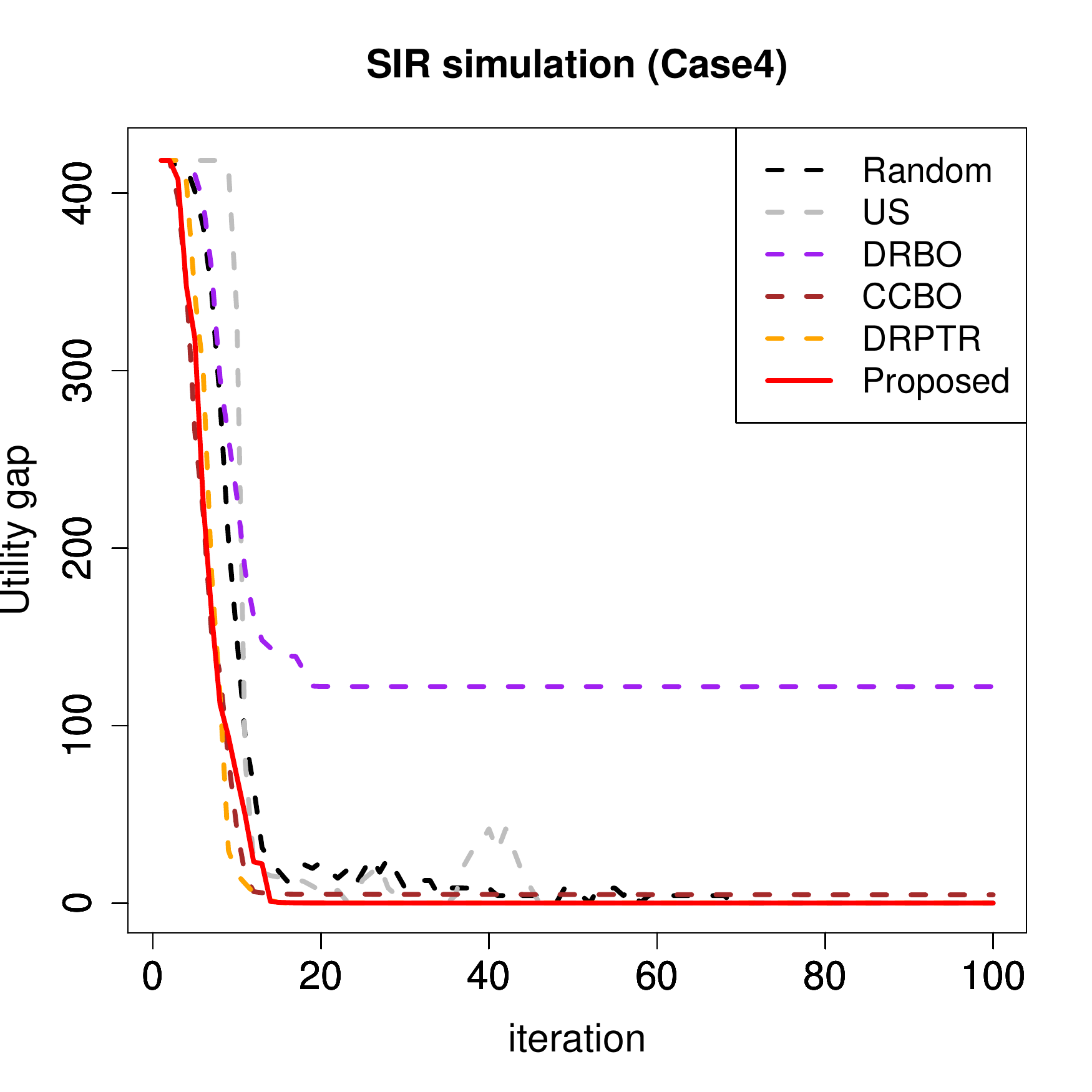} 
 \end{tabular}
\end{center}
 \caption{Average utility gap for SIR simulation experiments in the simulator setting.}
\label{fig:exp2}
\end{figure*}

\subsection{Infection Simulation}
We then applied the proposed method to the decision-making problem for simulation-based infectious diseases in the real world.
Here, we used the SIR model \cite{kermack1927contribution}, which is a commonly used model to describe the behavior of infection. 
The SIR model uses the contact rate $\beta\in [0,1]$ and the isolation rate $\gamma\in [0,1]$ to model the behavior of infection over time. 
In this experiment, we considered the grid points that divide the interval $[0.01,0.5]$ into 50 equal parts as $\beta$ and $\gamma$, and used them as input. 
Based on the SIR model, we defined the following two risk functions:
\begin{align*}
R_1 (\beta, \gamma ) &= n_ {\text{infected}} (\beta,\gamma ) -450 \beta + 800 \gamma -C_1, \\
R_2 (\beta, \gamma ) &= n_ {\text{infected}} (\beta,\gamma ) -C_2,
\end{align*}
where  $n_ {\text{infected}} (\beta,\gamma ) $, which is calculated by using the SIR model with $(\beta,\gamma)$,  
is the maximum number of infected within a given period. 
In addition, $C_1$ (resp. $C_2$) is a shift constant to match the absolute values of the maximum and minimum of $R_1 (\beta, \gamma )  $ (resp. $R_2 (\beta, \gamma )$).
While $R_2 (\beta, \gamma )$, which represents the number of infected people, is an intuitive risk function, $R_1 (\beta, \gamma )$ can be interpreted as an economic risk function. 
In fact, as the number of infected people increases, the economic risk increases. 
In addition, if the contact rate is large, that is, if freedom of action is not restricted, economic activity will not stagnate and the risk will be small.
On the other hand, if the isolation rate is large, economic activity will stagnate and the risk will increase.
Although these risk functions should be minimized, we multiplied them by minus one in our experiments in order to match the setting of this paper.
Also, $R_1 (\beta, \gamma ) $ can be interpreted as both an objective function and a constraint function, and the same is true for $R_2 (\beta, \gamma ) $. 
Similarly, the contact rate $\beta$ can be interpreted as both a design variable and an environmental variable, and the same is true for $\gamma$. 
For these reasons, we performed the following experiments:
\begin{description}
\item [Case1:] Design variable $x$: $\beta$, environmental variable $w$: $\gamma$, $f(x,w) = -R_1 (x,w)$, $g(x,w) = -R_2 (x,w)$.
\item [Case2:] Design variable $x$: $\beta$, environmental variable $w$: $\gamma$, $f(x,w) = -R_2 (x,w)$, $g(x,w) = -R_1 (x,w)$.
\item [Case3:] Design variable $x$: $\gamma$, environmental variable $w$: $\beta$, $f(x,w) = -R_1 (x,w)$, $g(x,w) = -R_2 (x,w)$.
\item [Case4:] Design variable $x$: $\gamma$, environmental variable $w$: $\beta$, $f(x,w) = -R_2 (x,w)$, $g(x,w) = -R_1 (x,w)$.
\end{description}
In all experiments, the simulator setting was considered, and $p^\ast_t (w) =1/50$ was used as the reference distribution.

With this setup, we took one initial point at random and ran the algorithms until the number of iterations reached 100.
The simulation was repeated 100 times and the average value of the utility gap at each iteration was calculated. 
From Figure \ref{fig:exp2}, it can be confirmed that the proposed method performs as well as or better than the comparison methods.

\section{Conclusion}
In this paper, we proposed the BO method for efficiently finding the optimal solution to the DRCC problem for the simulator and uncontrollable settings.
We showed that the proposed method can return an arbitrary accurate solution with high probability in a finite number of trials. 
Furthermore, through numerical experiments, we confirmed that the performance of the proposed method is superior to other comparison  methods.

\section*{Acknowledgement}
This work was partially supported by MEXT KAKENHI (21H03498, 20H00601, 17H04694, 16H06538), JSPS KAKENHI (JP21J14673), JST CREST (JPMJCR21D3), JST Moonshot R\&D (JPMJMS2033-05), JST AIP Acceleration Research (JPMJCR21U2), NEDO (JPNP18002, JPNP20006) and RIKEN Center for Advanced Intelligence Project.

\bibliography{myref}
\bibliographystyle{apalike}

\newpage
\section*{Appendix}

\setcounter{section}{0}
\renewcommand{\thesection}{\Alph{section}}
\renewcommand{\thesubsection}{\thesection.\arabic{subsection}}

\section{Proofs}
\subsection{Proof of Theorem \ref{thm:seido}}
From the proof of Theorem 4.1 in \cite{inatsu-2021}, with a probability of at least $1-3 \delta/4$ 
the following holds for any ${\bm x} \in \mathcal{X}$ and $t \geq 1$\footnote{They only consider the fixed candidate family $\mathcal{A}$, but the same argument also holds in the case of $\mathcal{A}_t \neq \mathcal{A}_{t^\prime} $. }:
$$
G_t ({\bm x} )  \leq u^{(G_t)}_t ({\bm x};\eta ),\quad G_t ({\bm x} ) \geq l^{(G_t)}_t ({\bm x};\eta ) -\xi.
$$
Here, if the stopping condition (S1) holds, then $ u^{(G_t)}_t ({\bm x};\eta ) \leq \alpha$ for any ${\bm x}$.
By combining this and 
$G_t ({\bm x} )  \leq u^{(G_t)}_t ({\bm x};\eta )$, we have $G_t({\bm x} )  \leq \alpha$.
This implies that the DRCC problem has no solution. 
On the other hand, if the stopping condition (S2) holds, 
$\hat{\bm x}_t $ satisfies that 
$$
l^{(G_t)}_t ({\bm x};\eta ) >\alpha -\xi.
$$
By using this and 
$G_t({\bm x} ) \geq l^{(G_t)}_t ({\bm x};\eta ) -\xi$, we obtain 
$G_t(\hat{\bm x} _t) \geq \alpha- 2 \xi$. 
Here, if the optimal solution ${\bm x}^\ast _t$ does not exist, from the definition it follows that $F_t ({\bm x}^\ast_t  ) -F_t (\hat{\bm x} _t) \leq 0 < \xi < 2 \xi $. 
Therefore, ${\bm x}^\ast_t $ is a $2 \xi$-accurate solution. 
Next, we consider the case where the optimal solution ${\bm x}^\ast _t$ exists.
From Lemma 5.1 in \cite{SrinivasGPUCB}, under the assumption on Theorem \ref{thm:seido}, with a probability of at least $1-\delta /4$ the following inequality holds for any 
$({\bm x},{\bm w} ) \in \mathcal{X} \times \Omega$ and $t \geq 1$: 
$$
f({\bm x} ,{\bm w} ) \in Q^{(f)}_t ({\bm x},{\bm w} ).
$$
Hence, it follows that $F_t ({\bm x}) \in Q^{(F_t)}_t ({\bm x})$. 
Moreover, because ${\bm x}^\ast_t$ satisfies $G_t({\bm x}^\ast _t )>\alpha$, then  
$G_t({\bm x}^\ast _t ) \leq u^{(G_t)}_t ({\bm x}^\ast_t ;\eta )$ 
with a probability of at least $1-3\delta/4$. 
Thus, 
we get 
${\bm x}^\ast _t \in H_t \cup M_t $. 
Hence, the following holds:
$$
F_t ({\bm x}^\ast _t ) \leq u^{(F_t)}_t ({\bm x}^\ast _t) \leq \max_{ {\bm x} \in H_t \cup M_t } u^{(F_t)}_t ({\bm x} ).
$$
Similarly, noting that $F_t (\hat{\bm x}_t ) \geq l^{(F_t)}_t (\hat{\bm x} _t ) = \max _{ {\bm x} \in H_t }  l^{(F_t)}_t ({\bm x}  )$, 
from the stopping condition (S2) it follows that 
$$
2 \xi > \xi > \max_{ {\bm x} \in H_t \cup M_t } u^{(F_t)}_t ({\bm x} ) -\max _{ {\bm x} \in H_t }  l^{(F_t)}_t ({\bm x}  )
\geq F_t ({\bm x}^\ast _t ) -F_t (\hat{\bm x}_t ).
$$
Therefore, $\hat{\bm x}_t$ is a $2\xi$-accurate solution.

\subsection{Proof of Theorem \ref{thm:convergence1}}
Let $T$ be the smallest positive integer satisfying \eqref{eq:mi_ineq}. 
Also let 
$({\bm x}_1,{\bm w}_1 ), \ldots , ({\bm x}_{T^2},{\bm w}_{T^2} ) $ be points selected by the algorithm.  
Here, one of the following holds for $T^2$:
\begin{description}
\item [Case1] There exists a positive integer $t \leq T^2$ such that $L_t = \mathcal{X} $.
  \item [Case2] For any positive integer $t \leq T^2$, $L_t \neq \mathcal{X} $.
\end{description}
If Case1 holds, then from the stopping condition (S1) the algorithm terminates. 
Next, we consider Case2.
Let 
$$
T_1 = \argmin _{ 1 \leq t \leq T}   \{  \sigma^{(f)2}_{t-1}  ({\bm x}_t,{\bm w}_t ) +  
 \sigma^{(g)2}_{t-1}  ({\bm x}_t,{\bm w}_t )  \}.
$$
Then, the following inequality holds:
\begin{align*}
T  \{  \sigma^{(f)2}_{{T}_1-1}  ({\bm x}_{{T_1}},{\bm w}_{{T_1} }) +  
 \sigma^{(g)2}_{{T_1}-1}  ({\bm x}_{{T_1}},{\bm w}_{{T_1}} )  \} &\leq \sum _{t=1}^T  \sigma ^{(f)2}_{t-1}  ({\bm x}_t , {\bm w}_t ) + \sum _{t=1}^T  \sigma ^{(g)2}_{t-1}  ({\bm x}_t , {\bm w}_t )  \\
& \leq C_{1,f}  \kappa^{(f)}_T +  C_{1,g} \kappa^{(g)}_T ,
\end{align*}
where the last inequality can be derived from Lemma 5.3 and 5.4 in  \cite{SrinivasGPUCB}. 
Thus, it follows that 
$$
\sigma^{(f)2}_{{T}_1-1}  ({\bm x}_{{T_1}},{\bm w}_{{T_1} }) +  
 \sigma^{(g)2}_{{T_1}-1}  ({\bm x}_{{T_1}},{\bm w}_{{T_1}} )  \leq \frac{C_{1,f}  \kappa^{(f)}_T +  C_{1,g} \kappa^{(g)}_T}{T}.
$$
In addition, from Definition \ref{def:w_selection_rule}, the following holds for any ${\bm w} \in \Omega $:
\begin{equation}
\sigma^{(g)2}_{{T_1}-1}  ({\bm x}_{{T_1}},{\bm w} ) \leq  \max_{ {\bm w} \in \Omega } (  \sigma^{(f)2}_{{T}_1-1}  ({\bm x}_{{T_1}},{\bm w}) +  
 \sigma^{(g)2}_{{T_1}-1}  ({\bm x}_{{T_1}},{\bm w} )   ) = \sigma^{(f)2}_{{T}_1-1}  ({\bm x}_{{T_1}},{\bm w}_{{T_1} }) +  
 \sigma^{(g)2}_{{T_1}-1}  ({\bm x}_{{T_1}},{\bm w}_{{T_1}} )  \leq \frac{C_{1,f}  \kappa^{(f)}_T +  C_{1,g} \kappa^{(g)}_T}{T}.
\nonumber 
\end{equation}
Hence, we have
$$
\beta_{g,T_1}  \sigma^{(g)2}_{{T_1}-1}  ({\bm x}_{{T_1}},{\bm w} ) \leq 
\frac{\beta_{g,T_1}  (C_{1,f}  \kappa^{(f)}_T +  C_{1,g} \kappa^{(g)}_T ) }{T} \leq 
\frac{\beta_{g,T^2}  (C_{1,f}  \kappa^{(f)}_T +  C_{1,g} \kappa^{(g)}_T ) }{T} .
$$
Here, from the theorem's assumption, it holds that 
$$
\frac{\beta_{g,T^2}  (C_{1,f}  \kappa^{(f)}_T +  C_{1,g} \kappa^{(g)}_T ) }{T}    < \eta^2/4.
$$
Therefore, we have $\beta^{1/2}_{g,T_1  } \sigma^{(g)} _{T_1-1 }  ({\bm x}_{T_1}  ,  {\bm w}  )
 < \eta /2$.
Furthermore, by combining this and Lemma A.3 in \cite{inatsu-2021}, we get $u^{(G_{T_1})}_{ {T}_1-1 }   ({\bm x}_{ {T}_1 };\eta) =
 l^{(G_{T_1})}_{ {T}_1-1 }   ({\bm x}_{ {T_1} };\eta) $.
Using this and the definition of $M_t$, it holds that ${\bm x}_{ {T_1} } \notin M_{ T_1} $.
Moreover, from Definition \ref{def:1}, it follows that ${\bm x}_{ {T_1} } \in H_{ T_1} \cup M_{T_1} $.
Thus, we have ${\bm x}_{ {T_1} } \in H_{ T_1}$. 
Similarly, we consider $({\bm x}_{T+1},{\bm w}_{T+1} ), \ldots ,  ({\bm x}_{2T},{\bm w}_{2T} )$. 
As with $T_1$, let 
$$
T_2 = \argmin _{ T+1 \leq t \leq 2T}   \{  \sigma^{(f)2}_{t-1}  ({\bm x}_t,{\bm w}_t ) +  
 \sigma^{(g)2}_{t-1}  ({\bm x}_t,{\bm w}_t )  \}.
$$
Then, the following inequality holds:
\begin{align*}
T  \{  \sigma^{(f)2}_{{T}_2-1}  ({\bm x}_{{T_2}},{\bm w}_{{T_2} }) +  
 \sigma^{(g)2}_{{T_2}-1}  ({\bm x}_{{T_2}},{\bm w}_{{T_2}} )  \} &\leq \sum _{t=T+1}^{2T}  \sigma ^{(f)2}_{t-1}  ({\bm x}_t , {\bm w}_t ) + \sum _{t=T+1}^{2T}  \sigma ^{(g)2}_{t-1}  ({\bm x}_t , {\bm w}_t ) .
\end{align*}
Furthermore, let $\sigma^{(f)2 }_0 ( {\bm x},{\bm w}  |  {\bm x}_{m:n},{\bm w}_{m:n} )$ be a posterior variance of $f({\bm x},{\bm w} )$ after adding $( {\bm x}_m,{\bm w}_m ),\ldots, 
( {\bm x}_n,{\bm w}_n )$. 
Then, it holds that
$$
\sum _{t=T+1}^{2T}  \sigma ^{(f)2}_{t-1}  ({\bm x}_t , {\bm w}_t ) 
\leq \sigma^{(f)2}_0  ({\bm x}_{T+1} , {\bm w}_{T+1} ) + \sum _{t=2}^T \sigma^{(f)2}_0 ( {\bm x}_{T+t} , {\bm w}_{T+t} | {\bm x}_{(T+1):(T+t-1)}, 
  {\bm w}_{(T+1):(T+t-1)}  ) \leq  C_{1,f}  \kappa^{(f)}_T.
$$
As with $ \sigma ^{(f)2}_{t-1}  ({\bm x}_t , {\bm w}_t ) $, the following holds for $ \sigma ^{(g)2}_{t-1}  ({\bm x}_t , {\bm w}_t ) $:
$$
\sum _{t=T+1}^{2T}  \sigma ^{(g)2}_{t-1}  ({\bm x}_t , {\bm w}_t ) 
\leq C_{1,g}  \kappa^{(g)}_T.
$$
Thus, the following inequality holds for $T_2$: 
\begin{align*}
T  \{  \sigma^{(f)2}_{{T}_2-1}  ({\bm x}_{{T_2}},{\bm w}_{{T_2} }) +  
 \sigma^{(g)2}_{{T_2}-1}  ({\bm x}_{{T_2}},{\bm w}_{{T_2}} )  \} \leq C_{1,f}  \kappa^{(f)}_T+C_{1,g}  \kappa^{(g)}_T.
\end{align*}
Hence, from the same argument as before, we obtain ${\bm x}_{T_2 } \in H_{T_2 }$.
By repeating this procedure up to $T^2$, we get the sequence ${\bm x}_{T_1}, {\bm x}_{T_2} , \ldots , {\bm x} _{T_T } $ satisfying ${\bm x}_{T_i} \in H_{T_i} $.

Next, from ${\bm x}_{T_i } \in H_{T_i } $, it follows that $a_{T_i-1}^{(G_{T_i-1} )} ({\bm x}_{T_i } ) =1 $ and 
$a_{T_i-1}^{(F_{T_i-1} )} ({\bm x}_{T_i } ) \leq  u^{(F_{T_i-1})} _{ T_i-1}  ({\bm x}_{T_i } ) - l^{(F_{T_i-1})} _{ T_i-1}  ({\bm x}_{T_i } )  $. 
Therefore, it holds that 
$$
a_{T_i-1} ({\bm x}_{T_i } ) \leq  u^{(F_{T_i-1})} _{ T_i-1}  ({\bm x}_{T_i } ) - l^{(F_{T_i-1})} _{ T_i-1}  ({\bm x}_{T_i } )  .
$$
Here, let $\tilde{p}({\bm w} ) \in \mathcal{A}_{ T_i-1} $ be a probability function satisfying 
$$
l^{(F_{T_i-1})} _{ T_i-1}  ({\bm x}_{T_i } ) = \sum_{ {\bm w} \in \Omega }   l^{(f)}_{  T_i-1}  ({\bm x}_{T_i },{\bm w} )  \tilde{p}({\bm w} ).
$$ 
Then, from the definition of $u^{(F_{T_i-1})} _{ T_i-1}  ({\bm x}_{T_i } ) $, the following holds:
$$
u^{(F_{T_i-1})} _{ T_i-1}  ({\bm x}_{T_i } ) \leq \sum_{ {\bm w} \in \Omega }   u^{(f)}_{  T_i-1}  ({\bm x}_{T_i },{\bm w} )  \tilde{p}({\bm w} ).
$$
Thus, we get
$$
 u^{(F_{T_i-1})} _{ T_i-1}  ({\bm x}_{T_i } ) - l^{(F_{T_i-1})} _{ T_i-1}  ({\bm x}_{T_i } )
 \leq 
\sum_{ {\bm w} \in \Omega } 2\beta^{1/2}_{f,T_i}  \sigma^{(f)}_{  T_i-1}  ({\bm x}_{T_i },{\bm w} )  \tilde{p}({\bm w} )
\leq 
2\beta^{1/2}_{f,T_i} \max_{  {\bm w} \in \Omega } \sigma^{(f)}_{  T_i-1}  ({\bm x}_{T_i },{\bm w} ).
$$
Hence, from Definition \ref{def:w_selection_rule} it follows that 
\begin{equation}
a^2_{T_i-1} ({\bm x}_{T_i } ) \leq 4 \beta_{f,T_i} (\max_{  {\bm w} \in \Omega } \sigma^{(f)}_{  T_i-1}  ({\bm x}_{T_i },{\bm w} ) )^2
\leq 4 \beta_{f,T^2} \max_{  {\bm w} \in \Omega } \sigma^{(f)2}_{  T_i-1}  ({\bm x}_{T_i },{\bm w} ) 
\leq
4 \beta_{f,T^2} ( \sigma^{(f)2}_{  T_i-1}  ({\bm x}_{T_i }, {\bm w}_{T_i } ) +  \sigma^{(g)2}_{  T_i-1}  ({\bm x}_{T_i }, {\bm w}_{T_i } ) ).
 \nonumber 
\end{equation}
Furthermore, let
$$
\tilde{T} = \argmin_{  T \in \{T_1,\ldots,T_T \}  }   a^2_{T-1} ({\bm x}_{T } ) .
$$
Then, the following inequality holds:
$$
T  a^2_{\tilde{T}-1} ({\bm x}_{\tilde{T} } ) \leq 4 \beta_{f,T^2} \sum_{i=1}^T   ( \sigma^{(f)2}_{  T_i-1}  ({\bm x}_{T_i }, {\bm w}_{T_i } ) +  \sigma^{(g)2}_{  T_i-1}  ({\bm x}_{T_i }, {\bm w}_{T_i } ) ) \leq 4 \beta_{f,T^2} (C_{1,f}  \kappa^{(f)}_T+C_{1,g}  \kappa^{(g)}_T ).
$$
This implies that
$$
 a^2_{\tilde{T}-1} ({\bm x}_{\tilde{T} } ) \leq \frac{ 4 \beta_{f,T^2} (C_{1,f}  \kappa^{(f)}_T+C_{1,g}  \kappa^{(g)}_T )  }{ T  }.
$$
On the other hand, for any ${\bm x} \in H_{\tilde{T}-1} \cup M_{ \tilde{T}-1 } $, 
from the definition of $a^{(G_{\tilde{T}-1 } ) } _{ \tilde{T}-1  }  ({\bm x} )$ it follows that 
$a^{(G_{\tilde{T}-1 } ) } _{ \tilde{T}-1  }  ({\bm x} ) \geq \xi$. 
Hence, $a_{ \tilde{T}-1  }  ({\bm x} )$ can be bounded as 
$$
(u^{(F_{\tilde{T}-1 })}_{\tilde{T}-1}   ({\bm x} ) - \max_{  {\bm x}  \in H_{  \tilde{T}-1 }  }  l^{(F_{\tilde{T}-1 })}_{\tilde{T}-1}   ({\bm x} )    ) \xi
=
(u^{(F_{\tilde{T}-1 })}_{\tilde{T}-1}   ({\bm x} ) - c^{( \text{best} ) }_{ \tilde{T}-1}     ) \xi
\leq
a_{ \tilde{T}-1  }  ({\bm x} ).
$$
Therefore, by using the theorem's assumption, we obtain
$$
(u^{(F_{\tilde{T}-1 })}_{\tilde{T}-1}   ({\bm x} ) - \max_{  {\bm x}  \in H_{  \tilde{T}-1 }  }  l^{(F_{\tilde{T}-1 })}_{\tilde{T}-1}   ({\bm x} )    ) ^2 
\leq 
\frac{ 4 \beta_{f,T^2} (C_{1,f}  \kappa^{(f)}_T+C_{1,g}  \kappa^{(g)}_T )  }{ T  } \xi^{-2} < \xi^2.
$$
Hence, ${\bm x} _{ \tilde{T} -1 } \in H_{ \tilde{T}-1 }$ and $
u^{(F_{\tilde{T}-1 })}_{\tilde{T}-1}   ({\bm x} ) - \max_{  {\bm x}  \in H_{  \tilde{T}-1 }  }  l^{(F_{\tilde{T}-1 })}_{\tilde{T}-1}   ({\bm x} )   
< \xi $ for any ${\bm x} \in H_{\tilde{T}-1} \cup M_{ \tilde{T}-1 } $.
Thus,  the stopping condition (S2) holds.

\subsection{Proof of Theorem \ref{thm:convergence2}}
The proof is almost the same as the proof of Theorem  \ref{thm:convergence1}.
Assume that there exists a positive integer $t \leq T^2$ such that $L_t = \mathcal{X}$.
Then, the stopping condition (S1) holds.

Next, we consider the case where $L_t \neq \mathcal{X}$ for any $t \leq T^2$. 
For each $i \in \{1,\ldots, T \}$, let  
$$
T_i = \argmin _{ (i-1)T+1 \leq t \leq iT}   \{  \mathbb{E} _{ {\bm w} } [ \sigma^{(f)2}_{t-1}  ({\bm x}_t,{\bm w} ) ] +  
 \mathbb{E} _{ {\bm w} } [ \sigma^{(g)2}_{t-1}  ({\bm x}_t,{\bm w} ) ] \}.
$$
Then, $T_1$ satisfies that 
\begin{align*}
T  \{ \mathbb{E}_{ {\bm w}} [ \sigma^{(f)2}_{{T}_1-1}  ({\bm x}_{{T_1}},{\bm w})] +  
 \mathbb{E}_{ {\bm w}} [ \sigma^{(g)2}_{{T}_1-1}  ({\bm x}_{{T_1}},{\bm w})]  \} &\leq \sum _{t=1}^T  \mathbb{E}_{  {\bm w} } [ \sigma ^{(f)2}_{t-1}  ({\bm x}_t , {\bm w} )] + \sum _{t=1}^T  \mathbb{E}_  {{\bm w} } [ \sigma ^{(g)2}_{t-1}  ({\bm x}_t , {\bm w} )]  .
\end{align*}
Furthermore, from Lemma 3 in \cite{kirschner2018information}, the following uniform bound holds with a probability of at least $1-\delta$:
\begin{align*}
\sum _{t=1}^T  \mathbb{E}_{  {\bm w} } [ \sigma ^{(f)2}_{t-1}  ({\bm x}_t , {\bm w} )] + \sum _{t=1}^T  \mathbb{E}_  {{\bm w} } [ \sigma ^{(g)2}_{t-1}  ({\bm x}_t , {\bm w} )] 
&\leq 
2\sum _{t=1}^T  \sigma ^{(f)2}_{t-1}  ({\bm x}_t , {\bm w}_t ) + 2 \sum _{t=1}^T  \sigma ^{(g)2}_{t-1}  ({\bm x}_t , {\bm w}_t ) + 16 \log (5/\delta) \\
& \leq 2 C_{1,f}  \kappa^{(f)}_T +  2 C_{1,g} \kappa^{(g)}_T + 16 \log (5/\delta) .
\end{align*}
By combining these, we have 
$$
 \mathbb{E}_{ {\bm w}} [ \sigma^{(f)2}_{{T}_1-1}  ({\bm x}_{{T_1}},{\bm w})] +  
 \mathbb{E}_{ {\bm w}} [ \sigma^{(g)2}_{{T}_1-1}  ({\bm x}_{{T_1}},{\bm w})]   \leq \frac{2 C_{1,f}  \kappa^{(f)}_T +  2 C_{1,g} \kappa^{(g)}_T + 16 \log (5/\delta)}{T}.
$$
In addition, noting that $p_{ \text{min} } >0$, the following inequality holds for any ${\bm w} \in \Omega$:
\begin{align*}
\sigma^{(g)2}_{{T_1}-1}  ({\bm x}_{{T_1}},{\bm w} ) &\leq  \max_{ {\bm w} \in \Omega } (  \sigma^{(f)2}_{{T}_1-1}  ({\bm x}_{{T_1}},{\bm w}) +  
 \sigma^{(g)2}_{{T_1}-1}  ({\bm x}_{{T_1}},{\bm w} )   ) \\
&\leq 
 p^{-1}_{\text{min}} ( \mathbb{E}_{ {\bm w}} [ \sigma^{(f)2}_{{T}_1-1}  ({\bm x}_{{T_1}},{\bm w})] +  
 \mathbb{E}_{ {\bm w}} [ \sigma^{(g)2}_{{T}_1-1}  ({\bm x}_{{T_1}},{\bm w})] )  \leq \frac{C_{2,f}  \kappa^{(f)}_T +  C_{2,g} \kappa^{(g)}_T + C_3}{T}.
\nonumber 
\end{align*}
Hence, we get 
$$
\beta_{g,T_1}  \sigma^{(g)2}_{{T_1}-1}  ({\bm x}_{{T_1}},{\bm w} ) \leq 
\frac{\beta_{g,T^2}  (C_{2,f}  \kappa^{(f)}_T +  C_{2,g} \kappa^{(g)}_T +C_3) }{T} .
$$
Moreover, from the theorem's assumption, it follows that 
$$
\frac{\beta_{g,T^2}  (C_{2,f}  \kappa^{(f)}_T +  C_{2,g} \kappa^{(g)}_T +C_3) }{T}    < \eta^2/4.
$$
Thus, by using the same argument as the proof of Theorem \ref{thm:convergence1}, we obtain ${\bm x}_{ {T_1} } \in H_{ T_1}$. 
By repeating this procedure up to $T^2$, we have the sequence ${\bm x}_{T_1}, {\bm x}_{T_2} , \ldots , {\bm x} _{T_T } $ satisfying ${\bm x}_{T_i} \in H_{T_i} $. 
From ${\bm x}_{T_i } \in H_{T_i } $, it follows that $a_{T_i-1}^{(G_{T_i-1} )} ({\bm x}_{T_i } ) =1 $ and 
$a_{T_i-1}^{(F_{T_i-1} )} ({\bm x}_{T_i } ) \leq  u^{(F_{T_i-1})} _{ T_i-1}  ({\bm x}_{T_i } ) - l^{(F_{T_i-1})} _{ T_i-1}  ({\bm x}_{T_i } )  $. 
Therefore, from the definition of the proposed AF, $a_{T_i-1} ({\bm x}_{T_i } ) $ can be bounded as
$$
a_{T_i-1} ({\bm x}_{T_i } ) \leq  u^{(F_{T_i-1})} _{ T_i-1}  ({\bm x}_{T_i } ) - l^{(F_{T_i-1})} _{ T_i-1}  ({\bm x}_{T_i } ) \leq 
2\beta^{1/2}_{f,T_i} \max_{  {\bm w} \in \Omega } \sigma^{(f)}_{  T_i-1}  ({\bm x}_{T_i },{\bm w} ).
$$
In addition, noting that $p_{ \text{min} } >0$ we get 
\begin{align*}
a^2_{T_i-1} ({\bm x}_{T_i } ) \leq 4 \beta_{f,T_i} (\max_{  {\bm w} \in \Omega } \sigma^{(f)}_{  T_i-1}  ({\bm x}_{T_i },{\bm w} ) )^2
&\leq 4 \beta_{f,T^2} \max_{  {\bm w} \in \Omega } \sigma^{(f)2}_{  T_i-1}  ({\bm x}_{T_i },{\bm w} ) \\
&\leq
4p^{-1}_{\text{min}}  \beta_{f,T^2} ( \mathbb{E}_{\bm w} [\sigma^{(f)2}_{  T_i-1}  ({\bm x}_{T_i }, {\bm w} )] +  \mathbb{E}_{\bm w} [\sigma^{(g)2}_{  T_i-1}  ({\bm x}_{T_i }, {\bm w} )] ).
\end{align*}
Let $\tilde{T} $ be an positive integer satisfying 
$$
\tilde{T} = \argmin_{  T \in \{T_1,\ldots,T_T \}  }   a^2_{T-1} ({\bm x}_{T } ) .
$$
Then, it follows that 
$$
T  a^2_{\tilde{T}-1} ({\bm x}_{\tilde{T} } ) \leq 4 p^{-1}_{\text{min} }\beta_{f,T^2} \sum_{i=1}^T    ( \mathbb{E}_{\bm w} [\sigma^{(f)2}_{  T_i-1}  ({\bm x}_{T_i }, {\bm w} )] +  \mathbb{E}_{\bm w} [\sigma^{(g)2}_{  T_i-1}  ({\bm x}_{T_i }, {\bm w} )] ) \leq 4  \beta_{f,T^2} (C_{2,f}  \kappa^{(f)}_T+C_{2,g}  \kappa^{(g)}_T +C_3 ).
$$
This implies that 
$$
 a^2_{\tilde{T}-1} ({\bm x}_{\tilde{T} } ) \leq \frac{ 4 \beta_{f,T^2} (C_{2,f}  \kappa^{(f)}_T+C_{2,g}  \kappa^{(g)}_T +C_3)  }{ T  }.
$$
Moreover, for any ${\bm x} \in H_{\tilde{T}-1} \cup M_{ \tilde{T}-1 } $, from the definition of  $a^{(G_{\tilde{T}-1 } ) } _{ \tilde{T}-1  }  ({\bm x} )$ it holds that $a^{(G_{\tilde{T}-1 } ) } _{ \tilde{T}-1  }  ({\bm x} ) \geq \xi$.
Hence, the following holds:
$$
(u^{(F_{\tilde{T}-1 })}_{\tilde{T}-1}   ({\bm x} ) - \max_{  {\bm x}  \in H_{  \tilde{T}-1 }  }  l^{(F_{\tilde{T}-1 })}_{\tilde{T}-1}   ({\bm x} )    ) \xi
=
(u^{(F_{\tilde{T}-1 })}_{\tilde{T}-1}   ({\bm x} ) - c^{( \text{best} ) }_{ \tilde{T}-1}     ) \xi
\leq
a_{ \tilde{T}-1  }  ({\bm x} ).
$$
Thus, from the theorem's assumption, it follows that  
$$
(u^{(F_{\tilde{T}-1 })}_{\tilde{T}-1}   ({\bm x} ) - \max_{  {\bm x}  \in H_{  \tilde{T}-1 }  }  l^{(F_{\tilde{T}-1 })}_{\tilde{T}-1}   ({\bm x} )    ) ^2 
\leq 
\frac{ 4 \beta_{f,T^2} (C_{2,f}  \kappa^{(f)}_T+C_{2,g}  \kappa^{(g)}_T+C_3 )  }{ T  } \xi^{-2} < \xi^2.
$$
Therefore, from  ${\bm x} _{ \tilde{T} -1 } \in H_{ \tilde{T}-1 }$ and 
$
u^{(F_{\tilde{T}-1 })}_{\tilde{T}-1}   ({\bm x} ) - \max_{  {\bm x}  \in H_{  \tilde{T}-1 }  }  l^{(F_{\tilde{T}-1 })}_{\tilde{T}-1}   ({\bm x} )   
< \xi $ 
for any ${\bm x} \in H_{\tilde{T}-1} \cup M_{ \tilde{T}-1 } $, the stopping condition (S2) holds.

\subsection{Proof of Theorem \ref{thm:seido-CC}}
Let $p^\ast_t ({\bm w} ) $ be an empirical distribution of ${\bm w}$. 
Then, from the Hoeffding's inequality, the following holds for any ${\bm w} \in \Omega $:
$$
\mathbb{P} ( |  p^\ast_t ({\bm w} ) - p^\dagger ({\bm w} ) | \geq \lambda ) \leq 2 \exp (-2t \lambda^2 ). 
$$
By letting 
$$
\lambda = \sqrt{  \frac{1}{2t} \log \left (   \frac{ |\Omega | \pi^2 t^2   }{3 \delta} \right )},
$$
with a probability of at least $1-\delta$, the following inequality holds for any $t \leq 1$ and ${\bm w} \in \Omega $:
$$
|  p^\ast_t ({\bm w} ) - p^\dagger ({\bm w} ) | \leq \lambda .
$$
Moreover, from the theorem's assumption, the distance between distributions can be expressed as 
$$
d(p^\ast_t ({\bm w} ), p^\dagger ({\bm w} )  ) = \sum _{ {\bm w} \in \Omega } |  p^\ast_t ({\bm w} ) - p^\dagger ({\bm w} ) |
 \leq |\Omega | \lambda = \epsilon_t .
$$
Thus, it follows that $p^\dagger ({\bm w} ) \in \mathcal{A}_t$. 
Here, if the stopping condition (S1) is satisfied, from Theorem \ref{thm:seido}, with a probability of at least $1-\delta$ the inequality $G_t ({\bm x} ) \leq  \alpha^\prime$ holds for any ${\bm x} \in \mathcal{X} $. 
Therefore, we get 
$$
\tilde{G} ({\bm x} ) = \{ \tilde{G} ({\bm x} )  -G_t ({\bm x} ) \} + G_t ({\bm x} ) \leq |\tilde{G} ({\bm x} )  -G_t ({\bm x} )| + \alpha^\prime.
$$ 
Furthermore, let $p_t ({\bm w} ) $ be a probability function satisfying 
$$
G_t ({\bm x} ) =   \sum_{ {\bm w} \in \Omega }  \1 [ g({\bm x},{\bm w} ) > h ] p_t ({\bm w} ).
$$
Then, noting that 
$p_t ({\bm w} ) , p^\dagger ({\bm w} ) \in \mathcal{A}_t$, $|\tilde{G} ({\bm x} )  -G_t ({\bm x} )|$ can be expressed as follows:
\begin{align*}
|\tilde{G} ({\bm x} )  -G_t ({\bm x} )|  &=\left | \sum_{ {\bm w} \in \Omega } \1 [ g({\bm x},{\bm w} ) > h ] \{p^\dagger ({\bm w}) -p_t ({\bm w} )
\}
\right | \\
&\leq 
\sum_{ {\bm w} \in \Omega } |p^\dagger ({\bm w}) -p_t ({\bm w} ) | \\
&=
\sum_{ {\bm w} \in \Omega } |p^\dagger ({\bm w}) -p^\ast_t ({\bm w} ) + p^\ast_t ({\bm w} )  -p_t ({\bm w} ) | \\
&\leq \sum_{ {\bm w} \in \Omega } |p^\dagger ({\bm w}) -p^\ast_t ({\bm w} )|   +   \sum_{ {\bm w} \in \Omega }    |  p^\ast_t ({\bm w} )  -p_t ({\bm w} ) | \\
&= d(p^\dagger ({\bm w}) , p^\ast_t ({\bm w} ) ) + d(p^\ast_t ({\bm w} ),  p_t ({\bm w} )  ) \leq 2 \epsilon_t \leq 2(1+\beta^{1/2}_{f,1} ) \epsilon_t < \xi.
\end{align*}
Hence, we have 
$$
\tilde{G} ({\bm x} ) \leq  |\tilde{G} ({\bm x} )  -G_t ({\bm x} )| + \alpha^\prime < \xi + \alpha^\prime = \xi + (\alpha -\xi) = \alpha .
$$
Thus, it holds that $\tilde{G} ({\bm x} )  < \alpha $ with a probability of at least $1-2 \delta$.
This implies that the CC problem has no solution.

Next, if the stopping condition (S2) is satisfied, $\hat{\bm x}_t$ satisfies the following inequality with a probability of at least $1-\delta$:
$$
G_t (\hat{\bm x}_t ) \geq \alpha^\prime -2 \xi = \alpha - 3 \xi.
$$
Noting that $p^\dagger ({\bm w} ) \in \mathcal{A}_t $, we obtain 
$$
\tilde{G}   (\hat{\bm x}_t ) \geq G_t (\hat{\bm x}_t ) \geq \alpha^\prime -2 \xi = \alpha - 3 \xi.
$$
Here, if the CC problem has no solution, then from the definition the following holds:
$$
\tilde{F} (\tilde{\bm x}^\ast ) - \tilde{F} (\hat{\bm x}_t) \leq 0 < 3 \xi.     
$$
Hence, $\hat{\bm x}_t$ is a $3 \xi$-accurate solution for the CC problem.
Similarly, if the optimal solution $\tilde{\bm x}^\ast $ to the CC problem exists, we get 
\begin{align}
&\tilde{F} (\tilde{\bm x}^\ast ) - \tilde{F} (\hat{\bm x}_t) \nonumber \\
&=\tilde{F} (\tilde{\bm x}^\ast ) - F_t (\tilde{\bm x}^\ast)  + F_t (\tilde{\bm x}^\ast) -F_t (\hat{\bm x}_t)  + F_t (\hat{\bm x}_t)   -  \tilde{F} (\hat{\bm x}_t). \label{eq:sep_all}
\end{align}
Because $\tilde{\bm x}^\ast $ is the optimal solution to the CC problem, we have $\tilde{G} (\tilde{\bm x}^\ast ) >\alpha $. 
Hence, by using this we obtain
\begin{align}
G_t  (\tilde{\bm x}^\ast ) &= \tilde{G} (\tilde{\bm x}^\ast ) +\{   G_t  (\tilde{\bm x}^\ast )        -\tilde{G} (\tilde{\bm x}^\ast ) \} \nonumber \\
& > \alpha -  |\tilde{G} ({\bm x} )  -G_t ({\bm x} )| > \alpha - \xi = \alpha ^\prime . \nonumber 
\end{align}
Therefore, from the definition of $ {\bm x}^\ast_t$, it follows that 
\begin{equation}
F_t  (\tilde{\bm x}^\ast ) \leq F_t ( {\bm x}^\ast_t ). \label{eq:sep_1}
\end{equation}
In addition, from $p^\dagger ({\bm w} ) \in \mathcal{A}_t$ and the definition of $F_t ({\bm x})$ and $\tilde{F} ({\bm x})$, the following inequality holds:
\begin{equation}
F_t (\hat{\bm x}_t)   -  \tilde{F} (\hat{\bm x}_t)  \leq 0. \label{eq:sep_2}
\end{equation}
Moreover, let 
$\tilde{p}_t ({\bm w} )  \in \mathcal{A}_t$ be a probability function satisfying 
$$
F_t (\tilde{\bm x}^\ast ) =   \sum_{ {\bm w} \in \Omega }   f(\tilde{\bm x}^\ast,{\bm w} )  \tilde{p}_t ({\bm w} ).
$$
Then, we get 
$$
| \tilde{F} (\tilde{\bm x}^\ast ) - F_t (\tilde{\bm x}^\ast) | 
\leq \sum_{ {\bm w} \in \Omega } | f( \tilde{\bm x}^\ast, {\bm w} ) |  | p^\dagger ({\bm w} ) - \tilde{p}_t ({\bm w} ) |.
$$
%
From Lemma 5.1 in \cite{SrinivasGPUCB}, the following holds with a probability of at least $1-\delta$:
$$
|  f( \tilde{\bm x}^\ast, {\bm w} ) | \leq \beta^{1/2}_{f,1} \sigma_0 ( \tilde{\bm x}^\ast, {\bm w} ) \leq \beta^{1/2}_{f,1}.
$$
By using this, we have 
\begin{align}
\tilde{F} (\tilde{\bm x}^\ast ) - F_t (\tilde{\bm x}^\ast) \leq &| \tilde{F} (\tilde{\bm x}^\ast ) - F_t (\tilde{\bm x}^\ast) |  \nonumber 
\\
&\leq  \beta ^{1/2} _{f,1} (  d(p^\dagger ({\bm w} ),p^\ast_t ({\bm w} ) )  +   d(p^\ast_t ({\bm w} ),\tilde{p}_t ({\bm w}) ) )
\leq 2 \beta^{1/2} _{f,1} \epsilon_t \leq 2(1+ \beta^{1/2}_{f,1} ) \epsilon_t < \xi. \label{eq:sep_3}
\end{align}
By substituting \eqref{eq:sep_1},\eqref{eq:sep_2} and \eqref{eq:sep_3} into \eqref{eq:sep_all}, we obtain 
$$
\tilde{F} (\tilde{\bm x}^\ast ) - \tilde{F} (\hat{\bm x}_t) < \xi  + F_t ({\bm x}^\ast _t ) - F_t (\hat{\bm x}_t ).
$$
Finally, from Theorem \ref{thm:seido}, noting that the ${\bm x}^\ast _t $ is a $2 \xi $-accurate solution for the DRCC problem, we get 
$$
F_t ({\bm x}^\ast _t ) - F_t (\hat{\bm x}_t ) < 2 \xi.
$$
Therefore, we get $\tilde{F} (\tilde{\bm x}^\ast ) - \tilde{F} (\hat{\bm x}_t) < 3 \xi $.

\section{Experimental Details}\label{app:exp_detail}
In this section, we give the details of the experiments conducted in Section \ref{sec:sec5}.

\paragraph{Experimental Parameter} 
The experimental parameters used in each experiment are given in Table \ref{tab:exp_para_syn} and \ref{tab:exp_para_SIR}.

\begin{table}[htb]
  \begin{center}
    \caption{Experimental parameters for each setting in the synthetic function}
\scalebox{0.90}{
    \begin{tabular}{c||c} \hline \hline
       & Parameter  \\ \hline 
   Simulator   &   \\
   Fixed   &   	  $\sigma^2_{f,\text{ker}} = 1, L_f=3, \sigma^2_{f,\text{noise}} =10^{-8}, \beta^{1/2}_{f,t}=3,      
                       \sigma^2_{g,\text{ker}} = 2500, L_g=4, \sigma^2_{g,\text{noise}} =10^{-4}, \beta^{1/2}_{g,t}=2, 
                       h=5, \alpha = 0.53$
 \\
   Data-driven   &   \\ \hline \hline
    \end{tabular}
}
    \label{tab:exp_para_syn}
  \end{center}
\end{table}

\begin{table}[htb]
  \begin{center}
    \caption{Experimental parameters for each setting in the SIR model simulation}
\scalebox{0.90}{
    \begin{tabular}{c||c} \hline \hline
       & Parameter  \\ \hline 
   Case1   &  $\sigma^2_{f,\text{ker}} = 5000, L_f=0.1, \sigma^2_{f,\text{noise}} =10^{-8}, \beta^{1/2}_{f,t}=3,      
                       \sigma^2_{g,\text{ker}} = 10^5, L_g=0.01, \sigma^2_{g,\text{noise}} =10^{-4}, \beta^{1/2}_{g,t}=2, 
                       h=320, \alpha = 0.85$ 
 \\ \hline
   Case2   &  $\sigma^2_{f,\text{ker}} = 5000, L_f=0.1, \sigma^2_{f,\text{noise}} =10^{-8}, \beta^{1/2}_{f,t}=3,      
                       \sigma^2_{g,\text{ker}} = 10^5, L_g=0.01, \sigma^2_{g,\text{noise}} =10^{-4}, \beta^{1/2}_{g,t}=2, 
                       h=320, \alpha = 0.85$ 
 \\ \hline 
 Case3   &   	  $\sigma^2_{f,\text{ker}} = 10^4, L_f=0.1, \sigma^2_{f,\text{noise}} =10^{-3}, \beta^{1/2}_{f,t}=2,      
                       \sigma^2_{g,\text{ker}} = 10^5, L_g=0.1, \sigma^2_{g,\text{noise}} =10^{-3}, \beta^{1/2}_{g,t}=3, 
                       h=100, \alpha = 0.69$
 \\ \hline 
 Case4   &   	  $\sigma^2_{f,\text{ker}} = 10^4, L_f=0.1, \sigma^2_{f,\text{noise}} =10^{-3}, \beta^{1/2}_{f,t}=2,      
                       \sigma^2_{g,\text{ker}} = 10^5, L_g=0.1, \sigma^2_{g,\text{noise}} =10^{-3}, \beta^{1/2}_{g,t}=3, 
                       h=100, \alpha = 0.69$
 \\ \hline  \hline
    \end{tabular}
}
    \label{tab:exp_para_SIR}
  \end{center}
\end{table}

\paragraph{True Distribution of Environmental Variables} 
We give the details of the true distribution $p^\dagger (w)$ considered in the uncontrollable setting in the synthetic function experiment.
Let $h (w; \mu,\sigma^2) $ be a probability density function of Normal distribution with mean $\mu$ and variance $\sigma^2$, and let 
$h (w) = 0.5 h (w;-5,10) + 0.5 h (w;5,10)$.  
Then, $p^\dagger (w)$ is given by 
$$
p^\dagger (w) = \frac{ h(w) }{  \sum _{w \in \Omega} h(w)} .
$$

\paragraph{DRPTR} 
The DRPTR AF is based on the expected classification improvement for $G_t ({\bm x} )$ after adding new data $({\bm x}^\ast,{\bm w}^\ast, y^{(g) \ast } )$. 
Let $l^{(G_t )}_t ( {\bm x};\eta |  {\bm x}^\ast,{\bm w}^\ast, y^{(g) \ast} )$ be a lower of the credible interval of $G_t ({\bm x} )$ at ${\bm x} $ after adding  $({\bm x}^\ast,{\bm w}^\ast, y^{(g) \ast } )$. 
Then, the expected classification improvement is given by 
\begin{align}
a_t ( {\bm x}^\ast,{\bm w}^\ast ) =    \sum _{  {\bm x} \in M_t }     \mathbb{E}_{ y^{(g) \ast } } [   \1[   l^{(G_t )}_t ( {\bm x};\eta |  {\bm x}^\ast,{\bm w}^\ast, y^{(g) \ast} ) >\alpha         ]         ].
\label{eq:original_DRPTR_AF}
\end{align}
In \cite{inatsu-2021}, they suggest combining \eqref{eq:original_DRPTR_AF} and RMILE AF proposed by \cite{zanette2018robust}.
The RMILE is based on the expected classification improvement for $g ({\bm x} ,{\bm w})$ after adding  $({\bm x}^\ast,{\bm w}^\ast, y^{(g) \ast } )$.
Let $l^{(g)}_t ( {\bm x},{\bm w} |  {\bm x}^\ast,{\bm w}^\ast, y^{(g) \ast} )$ be a lower of the credible interval of $g ({\bm x} ,{\bm w})$ at $ ( {\bm x}  ,{\bm w} )$ after adding  $({\bm x}^\ast,{\bm w}^\ast, y^{(g) \ast } )$. 
In our experiments, we used the following modified RMILE function:
\begin{align}
{\rm RMILE}_t ( {\bm x}^\ast,{\bm w}^\ast ) =    \sum _{ ( {\bm x},{\bm w} ) \in M_t \times \Omega }     \mathbb{E}_{ y^{(g) \ast } } [   \1[   l^{(g )}_t ( {\bm x},{\bm w} |  {\bm x}^\ast,{\bm w}^\ast, y^{(g) \ast} ) >h         ]         ].
\label{eq:modified_RMILE_AF}
\end{align}
Then, the DRPTR AF is defined as 
\begin{align}
{\rm DRPTR}_t ( {\bm x}^\ast,{\bm w}^\ast ) =  \max \{   a_t ( {\bm x}^\ast,{\bm w}^\ast ) , \gamma {\rm RMILE}_t ( {\bm x}^\ast,{\bm w}^\ast ) \}, \label{eq:used_DRPTR}
\end{align}
where $\gamma $ is a trade-off parameter. 
In all experiments, we set $\gamma =0.1$. 
From GP properties,  \eqref{eq:modified_RMILE_AF} can be calculated analytically \cite{zanette2018robust}.
In contrast, \eqref{eq:original_DRPTR_AF} can be represented in an exact form  (see, \cite{inatsu-2021}), but its computational cost is high.
In Lemma 3.3 in \cite{inatsu-2021}, an arbitrary-accurate approximation method for calculating \eqref{eq:original_DRPTR_AF} is proposed.
For all experiments, we used its lemma with approximation parameter $\zeta = 0.005 (|\Omega |+1 )$. 
This implies that the calculation error between the true \eqref{eq:original_DRPTR_AF} and approximated one is at most $\zeta$.

\paragraph{CCBO} 
The CCBO AF is based on the expected feasible improvement for the following CC problem:
$$
\max _{ {\bm x} \in \mathcal{X}} Z^{(F)}    ({\bm x} ) \quad \text{s.t.} \quad Z^{(G) }  ( ({\bm x} ) ) >\alpha,
$$
where $Z^{(F)}   ({\bm x} ) $ and $Z^{(G)}   ({\bm x} )$ are given by 
$$
Z^{(F)}   ({\bm x} ) = \sum _{ {\bm w} \in \Omega }   f({\bm x} ,{\bm w} )   p^\dagger ({\bm w} ) , 
Z^{(G)}   ({\bm x} ) = \sum _{ {\bm w} \in \Omega }   \1[ g ({\bm x} ,{\bm w} )>h]   p^\dagger ({\bm w} ) .
$$
In our experiments, to define  $Z^{(F)}   ({\bm x} ) $ and $Z^{(G)}   ({\bm x} )$, we used the reference distribution instead of 
$p^\dagger ({\bm w} )$. 
Let $Z^{(F)}  _t  ({\bm x} ) $ and $Z^{(G)}  _t ({\bm x} )$ be posterior distributions of  $Z^{(F)}   ({\bm x} ) $ and $Z^{(G)}   ({\bm x} )$, respectively.
 Here, the calculation of posterior distribution is based on GP posteriors of $f$ and $g$.
Then, the CCBO AF is given by 
\begin{align}
{\rm CCBO}_t  ({\bm x} ) =   \mathbb{E} [ \max \{   Z^{(F)}  _t  ({\bm x} ) - c^{(\text{feas} )}_t ,0 \} ]   \times \mathbb{P}  (Z^{(G)}   ({\bm x} ) > \alpha ), \label{eq:original_CCBO}
\end{align}
where $c^{(\text{feas} )}_t$ is given by 
\begin{align*}
c^{(\text{feas} )}_t= 
\left \{
\begin{array}{ll}
\max _{ {\bm x} \in S_t }    \mathbb{E} [    Z^{(F)}  _t  ({\bm x} ) ]   & \text{if} \ S_t \equiv \{ {\bm x} \in \mathcal{X} \mid \mathbb{E} [Z^{(G)}   ({\bm x} )]  > \alpha \} \neq \emptyset, \\
 \mathbb{E} [    Z^{(F)}  _t  (\tilde{\bm x} ) ]  , \ \tilde{\bm x} = \argmax_{ {\bm x} \in \mathcal{X} }   \mathbb{E} [Z^{(G)}   ({\bm x} )]  & \text{otherwise}
\end{array}
\right . .
\end{align*}
We select ${\bm x} _{t+1} $ by maximizing ${\rm CCBO}_t  ({\bm x} )$, that is, 
$$
{\bm x} _{t+1} = \argmax _{ {\bm x} \in \mathcal{X} } {\rm CCBO}_t  ({\bm x} ).
$$
In CCBO, the selection of ${\bm w} $ is based on the variance of ${\rm CCBO}_t  ({\bm x}_{t+1} )$ after adding $({\bm x}_{t+1},{\bm w}^\ast, y^{(g)\ast }  )$. 
Let ${\rm CCBO}_t  ({\bm x}_{t+1} |  {\bm x}_{t+1},{\bm w}^\ast, y^{(g)\ast } ) $ be a value of ${\rm CCBO}_t  ({\bm x}_{t+1} )$ after adding $({\bm x}_{t+1},{\bm w}^\ast, y^{(g)\ast }  )$. 
Then, we consider the variance of ${\rm CCBO}_t  ({\bm x}_{t+1} |  {\bm x}_{t+1},{\bm w}^\ast, y^{(g)\ast } ) $ with respect to $y^{(g)\ast }  $ :
\begin{align}
{\rm Var }_{ y^{(g)\ast }   } [ {\rm CCBO}_t  ({\bm x}_{t+1} |  {\bm x}_{t+1},{\bm w}^\ast, y^{(g)\ast } )  ] 
\equiv v_t ({\bm w} ^\ast ). \label{eq:one_step_EFI_var}
\end{align}
Using \eqref{eq:one_step_EFI_var} we select ${\bm w} _{t+1} $ as 
$$
{\bm w} _{t+1}   = \argmin _{  {\bm w}^\ast \in \Omega }   v_t ({\bm w} ^\ast ).
$$
Note that a part of the calculation of \eqref{eq:original_CCBO} and \eqref{eq:one_step_EFI_var} requires a Monte Carlo approximation, we took 1000 samples and approximated them.

\paragraph{SIR Model Simulation}
The SIR model is often used in infectious disease modeling and is given as the following differential equation using the contact rate $\beta$ and isolation rate $\gamma$:
\begin{align*}
\left \{
\begin{array}{l}
\frac{dS}{dT} = - \frac{\beta I S} {N} , \\
\frac{dI}{dT} =  \frac{\beta I S} {N} - \gamma I , \\
\frac{dR}{dT} = \gamma I ,
\end{array}
\right . 
\end{align*}
where $N = S + I + R$, and $S$, $I$ and $R$ are the number of susceptible, infected and removed people, respectively. 
In our experiment, we considered simulations from time $T=0$ to time $T=15$, and initial $S_0$, $I_0$, and $R_0$ were set to 990, 10, and 0, respectively.
By considering $dT \approx 0.005$ and discrete approximation of the differential equation, we calculated the number of $I_T $ for each 
$T \in \{ 0, 0.005, \ldots , 15 \} \equiv \mathcal{T} $.
Using this we defined the maximum number of infected people $n_{ \text{infected } } (\beta ,\gamma ) $ as 
$$
n_{ \text{infected } } (\beta ,\gamma )  = \max _{  T \in \mathcal{T} }   I_T .
$$

  \end{document}